\documentclass[a4paper,fleqn]{cas-dc}
\usepackage[numbers]{natbib}
\usepackage{subcaption}
\usepackage{algorithm}
\usepackage{amsmath,amsfonts,amsthm}
\usepackage{paralist}
\usepackage{multirow,multicol}
\usepackage{tabularx}
\usepackage{booktabs}
\usepackage{colortbl}
\usepackage{makecell}
\usepackage{color}
\usepackage{algorithm}         
\usepackage{algpseudocode}     
\usepackage{adjustbox}
\usepackage{subcaption}
\usepackage{ragged2e}
\makeatletter
\let\CAS@origRaggedRight\RaggedRight
\renewcommand\RaggedRight{\justifying}
\makeatother
\usepackage{microtype}
\def\tsc#1{\csdef{#1}{\textsc{\lowercase{#1}}\xspace}}
\tsc{WGM}
\tsc{QE}
\tsc{EP}
\tsc{PMS}
\tsc{BEC}
\tsc{DE}

\begin{document}
\let\WriteBookmarks\relax
\def\floatpagepagefraction{1}
\def\textpagefraction{.001}
\shorttitle{TAO-Net Network}
\shortauthors{Zihao Wang et~al.}

\title [mode = title]{TAO-Net: \underline{T}wo-stage \underline{A}daptive \underline{O}OD Classification \underline{Net}work for Fine-grained Encrypted Traffic Classification}

\tnotemark[1]

\tnotetext[1]{This work was supported by the National Key Research and Development Program of China (Grant No. 2022YFC2806603, ``Development of Marine Environmental Models and Disaster Early‑Warning Simulation Technologies for Deep‑Sea Mining Areas'').}

\author[1]{Zihao Wang}
\ead{250214020010@hhu.edu.cn}

\author[2]{Wei Peng}[ orcid=0000-0001-8179-1577]
\cormark[1]
\ead{pengwei@zgclab.edu.cn}

\author[1]{Junming Zhang}
\ead{20241807@hhu.edu.cn}

\author[3]{Jian Li}
\ead{jian263@sina.com}

\author[1]{Wenxin Fang}
\ead{250214020009@hhu.edu.cn}

\address[1]{College of Artificial Intelligence and Automation, Hohai University, China}
\address[2]{Zhongguancun Laboratory, Beijing, China}
\address[3]{College of Information Science and Engineering, Hohai University, China}

\cortext[cor1]{Corresponding author}

\begin{abstract}
Encrypted traffic classification aims to identify applications or services by analyzing network traffic data. One of the critical challenges is the continuous emergence of new applications, which generates Out-of-Distribution (OOD) traffic patterns that deviate from known categories and are not well represented by predefined models. Current approaches rely on predefined categories, which limits their effectiveness in handling unknown traffic types. Although some methods mitigate this limitation by simply classifying unknown traffic into a single ``Other" category, they fail to make a fine-grained classification. In this paper, we propose a \textbf{T}wo-stage \textbf{A}daptive \textbf{O}OD classification \textbf{Net}work (TAO-Net) that achieves accurate classification for both In-Distribution (ID) and OOD encrypted traffic. The method incorporates an innovative two-stage design: the first stage employs a hybrid OOD detection mechanism that integrates transformer-based inter-layer transformation smoothness and feature analysis to effectively distinguish between ID and OOD traffic, while the second stage leverages large language models with a novel semantic-enhanced prompt strategy to transform OOD traffic classification into a generation task, enabling flexible fine-grained classification without relying on predefined labels. Experiments on three datasets demonstrate that TAO-Net achieves 96.81-97.70\% macro-precision and 96.77-97.68\% macro-F1, outperforming previous methods that only reach 44.73-86.30\% macro-precision, particularly in identifying emerging network applications.
\end{abstract}



\begin{highlights}
\item TAO‑Net: two‑stage adaptive framework for fine‑grained encrypted‑traffic classification.
\item Stage‑1 hybrid detector fuses inter‑layer smoothness with PCA residuals for robust ID/OOD split.
\item Stage‑2 couples a transformer (ID) and an LLM steered by Semantic‑enhanced prompts (OOD) to label traffic.
\item Attains 96.8–97.7\% macro‑precision/F1 on CHNAPP, ISCXVPN and ISCXTor.
\item Precisely uncovers emerging applications, strengthening encrypted‑traffic monitoring and network security.
\end{highlights}

\begin{keywords}
Encrypted Traffic Classification \sep Out-of-Distribution Detection \sep Fine-grained Classification \sep Two-stage Framework \sep Large Language Models
\end{keywords}

\maketitle

\section{Introduction}
The rapid development of Internet technologies and the increasing diversification of network applications have led to unprecedented complexity in network traffic patterns, posing significant challenges to network management and security \cite{azab2024network, zhao2021network, afuwape2021performance}. Network traffic classification, especially for encrypted traffic, has become a cornerstone of modern network security, facilitating critical tasks such as intrusion detection, traffic anomaly analysis, and enforcement of security policies. 

\begin{figure}[t]
     \centering
    \includegraphics[width=0.5\textwidth]{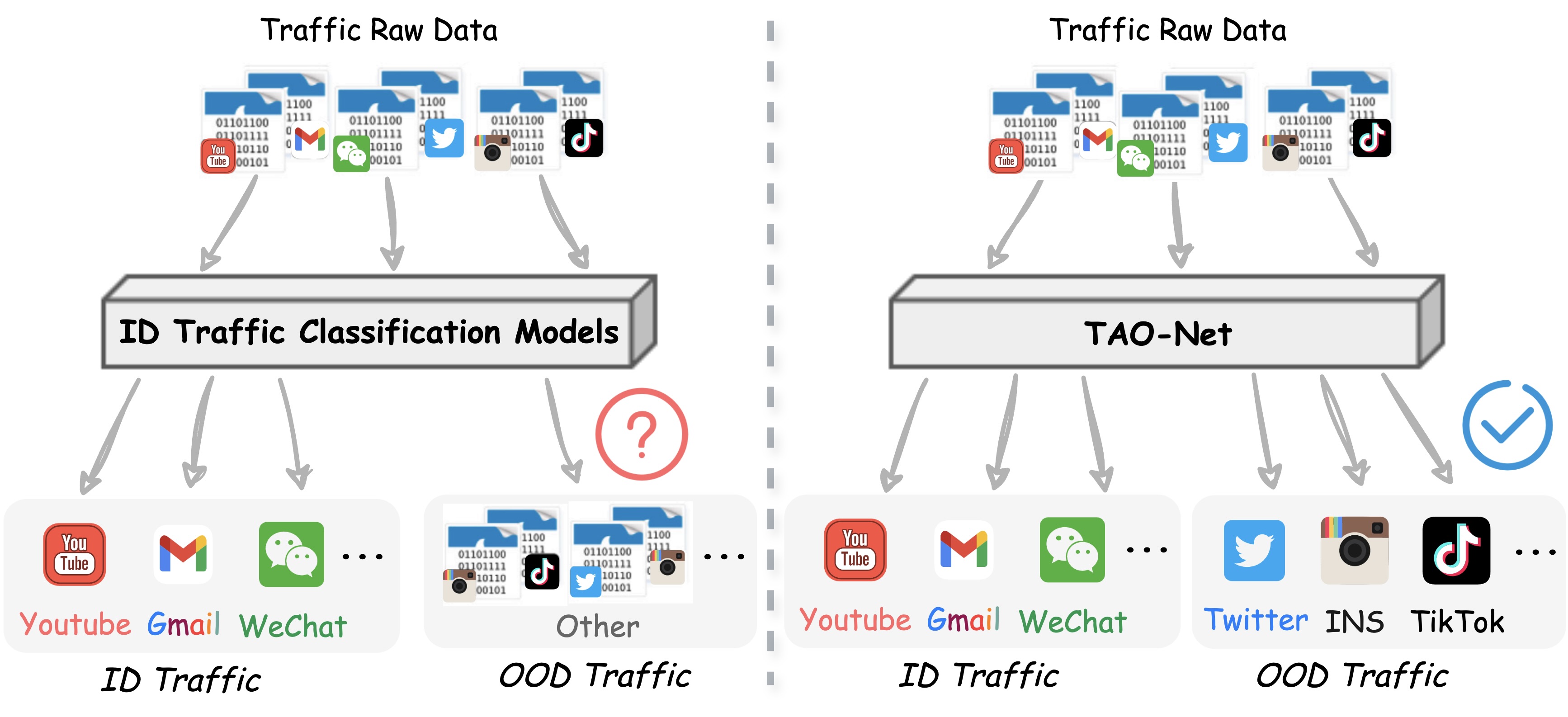}
    \caption{Comparison between previous ID traffic classification models (left) and our proposed TAO-Net (right). Previous models classify OOD traffic into a single ``Other'' category, while TAO-Net can identify specific applications in OOD traffic by leveraging LLMs' generation capabilities through a novel Semantic-enhanced Prompt Strategy (SPS), offering improved security monitoring.}
    \label{fig:overview}
\end{figure}

Current encrypted traffic classification methods primarily focus on In-Distribution (ID) data and show good performance for known categories. However, these approaches struggle with Out-of-Distribution (OOD) traffic generated by emerging applications, novel encryption methods, or protocol updates \cite{papadogiannaki2021survey, dong2024deep}. For example, novel P2P protocols masquerading as HTTPS connections, zero-day malware with custom encryption schemes, or emerging social media platforms with unique traffic patterns - all these new forms of traffic deviating significantly from previously observed patterns. This poses serious security risks, as previous classifiers cannot effectively identify these emerging traffic types, creating potential blind spots in network monitoring and security enforcement.

Although some existing approaches attempt to address OOD traffic by classifying all unknown patterns into a single ``Other'' category, this coarse-grained classification is inadequate for modern network security requirements \cite{zaki2022grain, tong2024method, miao2023spn}. Network administrators need precise identification of emerging applications and traffic patterns to effectively implement security policies, detect potential threats, and optimize network performance. For instance, when a new type of encrypted malware emerges or when legitimate applications adopt novel protocols, the ability to provide fine-grained classification of these unknown traffic patterns becomes crucial for maintaining network security. As illustrated in Fig. \ref{fig:overview}, previous models can only classify OOD traffic into a single ``Other'' category, limiting their practical utility in modern network environments \cite{miao2023spn,tong2024method,pathmaperuma2022deep}.

To address these challenges, we propose TAO-Net, a novel two-stage framework that transforms traffic classification into a generative task, enabling flexible and fine-grained identification of both known and unknown network traffic \cite{ring2019flow,iliyasu2019semi,wang2022two}. As shown in Fig. \ref{fig:overview}, unlike previous approaches, TAO-Net first employs a hybrid OOD detection mechanism that integrates transformer-based inter-layer transformation smoothness \cite{DBLP:conf/iclr/JelenicJTPS24} and feature analysis \cite{tong2024method} to accurately and reliably distinguish between ID and OOD traffic \cite{wu2022rtids,ullah2024ids,jorgensen2023extensible}. For OOD traffic, we leverage the powerful language understanding and generation capabilities of Large Language Models (LLMs) \cite{karanikolas2023large,abburi2023generative}, and combine them with our proposed innovative Semantic-enhanced Prompt Strategy (SPS) to generate specific application labels without relying on predefined categories. Furthermore, SPS designs three specialized modes (Strict, Complete, and Extended) to help narrow the generation space while maintaining flexibility, enabling the system to effectively handle complex real-world scenarios \cite{he2024sefd,koul2023prompt,cloutier2023fine}. In summary, this approach not only improves the precision of ID traffic classification but also can precisely identify unknown applications or services for effective threat detection and security monitoring.

The main contributions of this work are:
\begin{enumerate}
    \item \textbf{TAO-Net:} We propose a novel framework that enables fine-grained OOD encrypted traffic classification while maintaining high ID performance. Our innovative two-stage design adaptively processes ID and OOD traffic through separate specialized pathways, effectively addressing a critical gap in current approaches. To the best of our knowledge, this is the first comprehensive work for fine-grained OOD classification task in contemporary network environments.

    \item \textbf{Semantic-enhanced Prompt Strategy:} We transform traffic classification into a generative task using LLMs, introducing a novel three-level semantic-enhanced prompting strategy (SPS) that includes Strict, Complete, and Extended modes. This carefully designed prompt hierarchy effectively narrows the generation space while maintaining sufficient flexibility for complex real-world scenarios and diverse deployment conditions across different network environments.

    \item \textbf{Robust Security Performance:} Through comprehensive experiments on CHNAPP \cite{peng2024efficiently}, ISCXVPN \cite{gil2016characterization}, and ISCXTor \cite{lashkari2017characterization} datasets, we demonstrate that state-of-the-art (SOTA) encrypted traffic classification models \cite{meng2022packet,lin2022bert} show significant performance decline in OOD settings. The proposed method achieves remarkably high 96.81-97.70\% macro-precision in various encrypted traffic scenarios, representing a substantial 66.54-67.17\% relative improvement over baseline approaches (PacRep: 58.13-59.64\%). This significantly enhanced classification capability is particularly beneficial for security monitoring through fine-grained identification of emerging applications and services.
\end{enumerate}

\section{Related Work}
\subsection{Encrypted Traffic Classification Methods}
With the widespread adoption of network encryption technologies, encrypted traffic classification methods have evolved from traditional feature engineering to deep learning approaches. Early research primarily relies on manually designed traffic features. For instance, Moore et al. \cite{10.1145/1071690.1064220} propose utilizing statistical features of traffic for classification, including packet size and inter-arrival times. Zhang et al. \cite{zhang2014robust} further introduce a session behavior-based classification method by extracting behavioral features from traffic sessions. While these methods offer good interpretability, they involve cumbersome feature engineering processes and exhibit limited generalization capability.

The advancement of deep learning has brought new opportunities to encrypted traffic classification. Wang et al. \cite{wang2017end} pioneer the application of Convolutional Neural Networks (CNN) to encrypted traffic classification by transforming raw traffic data into image-like representations for feature learning. Lopez-Martin et al. \cite{lopez2017network} propose a Recurrent Neural Network (RNN)-based approach, enhancing classification performance through temporal feature modeling. Liu et al. \cite{liu2019fs} design FS-Net, a flow sequence network that combines CNN and LSTM to model traffic flows over time, improving performance across multiple datasets. Recently, attention mechanisms have further enhanced model performance \cite{peng2024efficiently,reza2022multi}, as exemplified by Reza et al. \cite{reza2022multi}, who propose a multi-head attention-based Transformer model to capture long-term dependencies in traffic data. Peng et al. \cite{wei2024bottom} propose a simple yet highly effective Aggregator and Separator Network (ASNet) for encrypted traffic understanding, which significantly enhances model performance by leveraging a two-stage processing approach.

However, these methods primarily focus on classifying ID data, with limited capability in handling OOD data. Some studies have attempted to address OOD traffic by adding an "unknown" category, such as the SPN framework proposed by Miao et al. \cite{miao2023spn}, but these approaches cannot achieve fine-grained recognition of unknown categories.

\subsection{OOD Detection Methods}
OOD detection and classification represents a significant research direction in machine learning. In general domains, early OOD detection primarily relies on statistical methods and distance metrics. Hendrycks and Gimpel \cite{hendrycks2016baseline} first propose using the maximum softmax prediction probability as a baseline method for OOD detection. Subsequently, Liang et al. \cite{DBLP:conf/iclr/LiangLS18} introduce the ODIN method, enhancing OOD detection performance through temperature scaling and input preprocessing. Lee et al. \cite{lee2018simple} develop a Mahalanobis distance-based detection method, identifying OOD samples through class-conditional Gaussian distributions.

In the network traffic domain, OOD detection faces significant challenges due to the flexible nature of network applications. Wu et al. \cite{wu2023gnnsafe} pioneer the application of OOD detection to network traffic analysis by proposing a deep generative model-based method. Liang et al. \cite{liang2024defending} design an adaptive threshold mechanism, enhancing OOD detection robustness through dynamic decision boundary adjustment. More recent innovations include the work of Jelenic et al. \cite{DBLP:conf/iclr/JelenicJTPS24}, who introduce an OOD detection method based on transformer inter-layer transformation smoothness, achieving effective detection without training data. Tong et al. \cite{tong2024method} propose a feature analysis-based method specifically for encrypted mobile traffic classification scenarios, quantifying sample deviation from training distribution through principal component analysis and residual feature vectors.

Considering the smoothness characteristics of inter-layer feature transformations in deep neural networks and the structural properties of traffic data, we adopt a detection strategy that integrates transformer inter-layer behavior and feature analysis. This method effectively captures the differences between ID and OOD data while reducing computational overhead without compromising detection accuracy.

\subsection{Generative Models in Classification Tasks}
The application of generative models in classification tasks has evolved significantly over recent years from traditional generative models and pre-trained language models to large language models. Early research focuses primarily on Generative Adversarial Networks (GANs) and Variational Autoencoders (VAEs). Salimans et al. \cite{salimans2016improved} pioneer the exploration of using GANs for semi-supervised classification tasks. Kumar et al. \cite{kumar2017semi} propose a VAE-based classification framework that enhances classification performance by learning latent representations of data.

The development of pre-trained language models and LLMs has introduced new paradigms for classification tasks. BERT \cite{devlin2018bert} achieves breakthrough progress in multiple classification tasks through its bidirectional encoder architecture and masked language modeling task. BART \cite{lewis2019bart} and T5 \cite{raffel2020exploring} further unify the transformation from classification to generation tasks through sequence-to-sequence architectures, providing new perspectives for handling unknown categories. Recent advancements in LLMs have provided more powerful solutions for classification tasks. These models (such as GPT-4, LLaMA and ChatGLM \cite{ouyang2022training,achiam2023gpt,touvron2023llama,glm2024chatglm}) have demonstrated exceptional generalization capabilities, effectively transforming classification tasks into generation tasks through instruction tuning. Through Reinforcement Learning from Human Feedback (RLHF), these models have demonstrated robust reasoning abilities and knowledge transfer capabilities, particularly excelling in cross-domain scenarios.

Compared to previous methods, our proposed TAO-Net framework offers significant advantages: First, through its two-stage adaptive design, it achieves different processing of ID and OOD data, ensuring classification performance for known categories while enhancing recognition capabilities for unknown categories. Second, it innovatively introduces LLMs into the network traffic classification domain, fully utilizing their powerful cross-domain understanding and generation capabilities to achieve fine-grained recognition of unknown categories. Finally, through the design of our Semantic-enhanced prompt strategy (SPS) as a prior knowledge injection mechanism, it effectively narrows the model's search space, improving the accuracy and interpretability of classification results.

\begin{table}[t]
\caption{Notation and Description}
\label{tab:notation}
\begin{tabular}{p{0.35\columnwidth}p{0.55\columnwidth}}
\toprule
\textbf{Notation} & \textbf{Description} \\
\midrule
$M = \{X_1, X_2, ..., X_N\}$ 
  & A set of $N$ encrypted traffic samples \\
$X_i = \{x_{i1}, x_{i2}, ..., x_{ij}\}$ 
  & A sequence of tokens in the i-th traffic sample \\
$x_{ij}$ 
  & The j-th token in the i-th traffic sample \\
$Y_{ID}$ 
  & ID label set \\
$Y_{OOD}$ 
  & OOD label set \\
$\phi_{ID}: M \rightarrow Y_{ID}$ 
  & Mapping function for ID classification \\
$\phi_{OOD}: M \rightarrow Y_{OOD}$ 
  & Mapping function for OOD detection \\
$D_{ID}$ 
  & ID classification dataset \\
$D_{OOD}$ 
  & OOD detection dataset \\
$\theta_{det}$ 
  & Parameter set for OOD detection \\
$\theta_{cls}$ 
  & Parameter set for ID classification \\
$\delta$ 
  & Decision threshold for OOD detection \\
$S(X_i)$ 
  & OOD score for sample $X_i$ \\
$T = \{T_1, T_2, T_3\}$ 
  & Set of prompt learning templates \\
$P(y|X_i)$ 
  & Classification probability distribution for sample $X_i$ \\
\bottomrule
\end{tabular}
\end{table}

\section{Preliminaries}
\subsection{Notation Definition}
To facilitate the subsequent technical discussions and promote better understanding, we first introduce the key notations used throughout this paper, as shown in Table \ref{tab:notation}. These notations form the foundation for describing our proposed TAO-Net framework and its various components, and serve as concise mathematical references for readers.

\subsection{Problem Definition}
As shown in Table \ref{tab:notation}, given an encrypted network traffic packet $X_i$ with its corresponding token sequence $\{x_{i1}, x_{i2}, ..., x_{ij}\}$, our goal is to assign an appropriate label $y$ from either $Y_{ID}$ or $Y_{OOD}$. The objective of encrypted traffic classification is to learn a mapping function $\phi\!: M \rightarrow Y$ that maps encrypted traffic samples to a predefined set of categories. However, real-world applications require handling encrypted traffic data from both ID and OOD sources simultaneously \textit{in diverse practical deployments}. Therefore, we formalize this problem as learning two distinct mapping functions:
1). ID Classification Mapping: $\phi_{ID}$: $M \rightarrow Y_{ID}$
2). OOD Classification Mapping: $\phi_{OOD}$: $M \rightarrow Y_{OOD}$
where $Y_{ID}$ represents the predefined set of known categories, and $Y_{OOD}$ denotes an open category space that requires dynamic generation of fine-grained category labels through advanced generative models.

\subsection{Basic Concepts} \label{sec:basic-concepts}
Several key concepts are essential for understanding our methodology:

\noindent\textbf{Encrypted Traffic.} Network communication data that has undergone encryption processing. In encrypted traffic raw data, for example, the packet structure includes network layer information (IP version, packet length, TTL), transport layer information (TCP sequence number, acknowledgment number, flags), and encrypted payload. Due to encryption, the actual payload content cannot be directly accessed, making traffic classification primarily dependent on protocol header information and statistical features of the encrypted payload.

\noindent\textbf{ID/OOD Data.} ID data refers to encrypted traffic generated by applications present during the training phase, with feature distributions similar to the training data. OOD data originates from previously unseen applications, exhibiting significant differences in feature distributions from the training data. For example, when a model is trained solely on encrypted traffic from traditional social applications, traffic from new instant messaging applications during testing constitutes OOD data.

\noindent\textbf{Zero-shot Classification.} The model's ability to recognize previously unseen categories without requiring additional training. This capability is crucial in encrypted traffic classification due to the continuous evolution of network applications. Our approach transforms the classification task into a generation task, leveraging the cross-domain understanding capabilities of LLMs to achieve zero-shot classification of OOD encrypted traffic data without requiring labeled data for emerging application categories.

\section{Proposed Method}
\label{sec:proposed_method}

In this section, we present the proposed TAO-Net (\textbf{T}wo-stage \textbf{A}daptive \textbf{O}OD classification \textbf{Net}work), which enables adaptive classification of encrypted network traffic through an innovative two-stage design. Specifically, we introduce two parameter sets: \(\theta_{\det}\) for OOD detection and \(\theta_{\text{cls}}\) for ID traffic classification. Unlike previous methods that only focus on known classes, TAO-Net not only identifies ID traffic with high accuracy but also supports fine-grained classification of OOD traffic by leveraging the generation capabilities of LLMs and proposed SPS.

\subsection{Overview of TAO-Net}
\label{sec:overview_taonet}
\begin{figure*}
\centering
\includegraphics[width=0.99\textwidth]{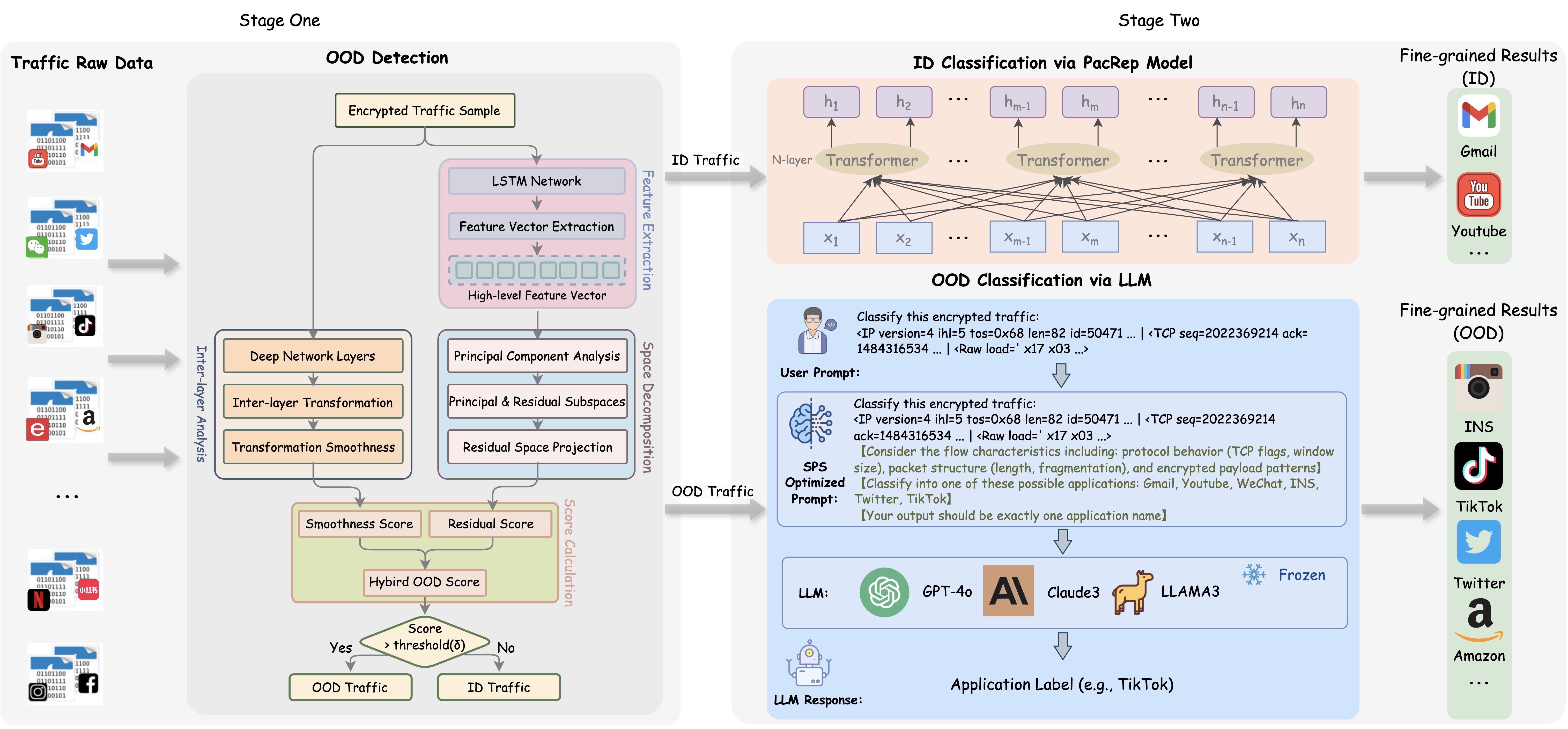} 
\caption{Framework of TAO-Net. Stage one performs OOD detection through a hybrid mechanism that integrates inter-layer transformation smoothness and feature analysis. Stage two conducts adaptive classification: ID traffic follows a transformer-based classification path for known categories, while OOD traffic is processed through an LLM with SPS, enabling fine-grained identification of emerging applications.}
\label{fig:framework} 
\end{figure*}

As illustrated in Fig.~\ref{fig:framework}, TAO-Net employs a two-stage architecture for encrypted traffic classification. Stage one performs OOD detection to determine whether incoming traffic belongs to ID or OOD categories. Stage two conducts adaptive classification based on the detection results, applying different strategies for ID and OOD traffic respectively.

In the first stage, TAO-Net leverages a hybrid OOD detection mechanism that combines inter-layer transformation smoothness and feature-based anomaly detection. This stage, parameterized by \(\theta_{\det}\), effectively distinguishes between ID and OOD traffic by analyzing both feature distributions and deep network behaviors.

The second stage implements adaptive classification strategies based on the detection results:

\begin{itemize}
\item[1)] For traffic detected as ID, the framework employs a transformer-based classifier (parameterized by \(\theta_{\text{cls}}\)) to achieve precise classification of known categories through deep representation learning.

\item[2)] For traffic detected as OOD, TAO-Net leverages an LLM and transforms the classification problem into a generation task. Combined with our Semantic-enhanced Prompt Strategy (SPS), the model can produce fine-grained class labels for unknown categories without relying on predefined lists.
\end{itemize}

The detailed workflow is summarized in Algorithm~\ref{alg:taonet}.

\begin{algorithm}[t]
\caption{TAO-Net: Two-stage Adaptive OOD Classification Framework}
\label{alg:taonet}
\begin{algorithmic}[1]
\Require 
    \Statex Encrypted traffic sample \(X \in \mathbb{R}^{j \times d}\).
    \Statex Training set of ID samples \(D_{ID} = \{X_1, ..., X_N\}\).
    \Statex Parameters for OOD detection \(\theta_{\det}\), parameters for ID classification \(\theta_{\text{cls}}\).
    \Statex Thresholds \(\gamma, \alpha, \delta\).
    \Statex Prompt template set \(T = \{T_1, T_2, T_3\}\).

\Ensure 
    \Statex Predicted traffic label \(y\).

\vspace{0.5em}
\textbf{Stage 1: OOD Detection}
\State \(\{\phi_i\}_{i=1}^N \leftarrow \mathrm{LSTM}_{\theta_{\det}}(D_{ID})\) \Comment{Feature extraction for ID data}
\State \(\mu, \sigma \leftarrow \mathrm{ComputeStatistics}(\{\phi_i\}_{i=1}^N)\)
\State \(C \leftarrow \mathrm{ComputeCov}\bigl(\{(\phi_i-\mu)/\sigma\}_{i=1}^N\bigr)\)
\State \(\{\lambda_i, v_i\} \leftarrow \mathrm{EigenDecomp}(C)\)
\State \(k \leftarrow \mathrm{DetermineK}(\{\lambda_i\}, \gamma)\)
\State \(P_{\mathcal{R}} \leftarrow \mathrm{ConstructProjector}(\{v_i\}_{i=k+1}^m)\)
\State \(\phi_{\mathrm{test}} \leftarrow \bigl(\mathrm{LSTM}_{\theta_{\det}}(X)-\mu\bigr)/\sigma\)
\State \(s_1 \leftarrow \|P_{\mathcal{R}} \cdot \phi_{\mathrm{test}}\|_2\)  \Comment{Residual score}
\State \(s_2 \leftarrow \sum_{l=1}^L \| F_l(X) - F_{l-1}(X)\|_2\) \Comment{Smoothness score}
\State \(S \leftarrow \alpha\,s_1 + (1-\alpha)\,s_2\) \Comment{Hybrid OOD score}
\State \(\mathrm{is\_ood} \leftarrow (S > \delta)\)

\vspace{0.5em}
\textbf{Stage 2: Adaptive Classification}
\If{not \(\mathrm{is\_ood}\)} \Comment{ID Classification Branch}
    \State \(H, v \leftarrow \mathrm{Transformer}_{\theta_{\text{cls}}}(X)\)
    \State \(A \leftarrow \mathrm{softmax}\Bigl(\frac{QK^T}{\sqrt{d_k}}\Bigr) V\)
    \State \(y \leftarrow \mathrm{softmax}(W \cdot A + b)\)
\Else \Comment{OOD Classification Branch}
    \State \(Q \leftarrow \mathrm{StandardizeFeatures}(X)\)
    \State \(p \leftarrow \mathrm{ConstructPrompt}(Q, T)\)
    \State \(y \leftarrow \arg\max_{y'\in Y_{OOD}}\, P(y' \mid X)\)
\EndIf
\State \Return \(y\)
\end{algorithmic}
\end{algorithm}

\subsection{OOD Detection Mechanism}
\label{sec:ood_detection}
The first stage of TAO-Net is OOD detection, parameterized by \(\theta_{\det}\). It leverages two key insights from recent studies: \emph{(i)} ID data typically exhibits smoother inter-layer feature transformations in deep networks~\cite{DBLP:conf/iclr/JelenicJTPS24}, and \emph{(ii)} ID data clusters within lower-dimensional manifolds in a suitable feature space~\cite{tong2024method}. Below, we describe how these insights are integrated into a unified hybrid scoring mechanism.

\paragraph{Feature Extraction.} 
Given an encrypted traffic sample \(X_i = \{x_{i1}, x_{i2}, \dots, x_{ij}\} \in \mathbb{R}^{j \times d}\), we employ an LSTM (with parameters \(\theta_{\det}\)) to extract a high-level feature vector \(\phi(X_i)\). The LSTM updates are computed as:
\begin{align}
i_t &= \sigma(\mathbf{W}_i [h_{t-1}, x_{ij}] + \mathbf{b}_i), \\
f_t &= \sigma(\mathbf{W}_f [h_{t-1}, x_{ij}] + \mathbf{b}_f), \\
o_t &= \sigma(\mathbf{W}_o [h_{t-1}, x_{ij}] + \mathbf{b}_o), \\
\tilde{c}_t &= \tanh(\mathbf{W}_c [h_{t-1}, x_{ij}] + \mathbf{b}_c), \\
c_t &= f_t \odot c_{t-1} + i_t \odot \tilde{c}_t, \\
h_t &= o_t \odot \tanh(c_t),
\end{align}
where \(i_t, f_t, o_t\) denote input, forget, and output gates, respectively, \(c_t\) is the cell state, and \(h_t \in \mathbb{R}^d\) is the hidden state at time \(t\). After processing the entire sequence (\(t=j\)), we define
\begin{equation}
\phi(X_i) \;=\; h_j,
\end{equation}
where \(h_j\) is the final hidden state, serving as the extracted feature representation of \(X_i\).

\paragraph{Space Decomposition.} 
We further decompose the feature space \(\mathbb{R}^d\) into two orthogonal subspaces using Principal Component Analysis (PCA). First, we compute the covariance matrix:
\begin{equation}
\label{eq:pca_cov}
\mathbf{C} \;=\; \frac{1}{N}\,\sum_{i=1}^N \Bigl(\phi(X_i) - \boldsymbol{\mu}\Bigr)\,\Bigl(\phi(X_i) - \boldsymbol{\mu}\Bigr)^T,
\end{equation}
where \(N\) is the number of ID training samples, \(\phi(X_i)\) is the LSTM-extracted feature for the \(i\)-th sample, and \(\boldsymbol{\mu}\) is the mean vector of all \(\{\phi(X_i)\}\). Eigen-decomposition of \(\mathbf{C}\) yields eigenvectors \(\{\mathbf{v}_i\}_{i=1}^m\) with corresponding eigenvalues \(\{\lambda_i\}_{i=1}^m\). We choose the top \(k\) principal components to form the principal subspace \(\mathcal{P}\), while the remaining \(m-k\) components form the residual subspace \(\mathcal{R}\):
\begin{align}
\label{eq:pca_decompose_p}
\mathcal{P} \;&=\; \mathrm{span}\{\mathbf{v}_1, \dots, \mathbf{v}_k\},\\
\label{eq:pca_decompose_r}
\mathcal{R} \;&=\; \mathrm{span}\{\mathbf{v}_{k+1}, \dots, \mathbf{v}_m\}.
\end{align}
We determine \(k\) by a cumulative variance ratio \(\gamma \in (0,1]\):
\begin{equation}
\label{eq:choose_k}
k \;=\; \min\Bigl\{m \,\Bigm|\; \frac{\sum_{i=1}^m \lambda_i}{\sum_{j=1}^m \lambda_j} \;\ge\; \gamma\Bigr\},
\end{equation}
where \(\lambda_i\) is the \(i\)-th largest eigenvalue of \(\mathbf{C}\), and \(m\) is the total number of eigenvalues.

\paragraph{Hybrid OOD Score Calculation.}
To decide whether a test sample \(X_i\) is OOD, we compute a hybrid score \(S(X_i)\) that integrates residual space projection and inter-layer smoothness:
\begin{equation}
\label{eq:hybrid_score}
S(X_i) \;=\; \alpha\,\Bigl\|P_{\mathcal{R}}\,\phi(X_i)\Bigr\|_2 \;+\; \bigl(1-\alpha\bigr)\,\sum_{l=1}^L \bigl\|F_l(X_i) - F_{l-1}(X_i)\bigr\|_2,
\end{equation}
where \(P_{\mathcal{R}}\) is the projection operator onto the residual subspace \(\mathcal{R}\), \(\phi(X_i)\) is the LSTM-extracted feature, \(F_l\) denotes the feature representation at layer \(l\) of a deep neural network, and \(\alpha \in [0,1]\) is a balancing factor. If \(S(X_i) > \delta\), we regard \(X_i\) as OOD; otherwise, it is classified as ID.

\subsection{ID Traffic Classification}
\label{sec:id_classification}
If a sample is determined to be ID, TAO-Net adopts a transformer-based classifier \cite{meng2022packet} parameterized by \(\theta_{\text{cls}}\). For each ID sample \(X_i\), we compute:
\begin{equation}
H_i, v_i \;=\; \mathrm{Transformer}_{\theta_{\text{cls}}}(X_i),
\end{equation}
where \(H_i\) represents layer-wise hidden states, and \(v_i\) is the final output embedding (or classification token representation). An attention mechanism is then applied:
\begin{equation}
A \;=\; \mathrm{softmax}\Bigl(\frac{Q K^T}{\sqrt{d_k}}\Bigr)\,V,
\end{equation}
where \(Q, K, V\) are query, key, and value matrices derived from \(H_i\). Finally, the model generates classification probabilities:
\begin{equation}
P(y \mid X_i) \;=\; \mathrm{softmax}\bigl(W \cdot A + b\bigr),
\end{equation}
where \(W\) and \(b\) are learnable parameters. This method effectively captures the hierarchical dependencies in encrypted traffic to achieve high accuracy on known categories.

\subsection{LLMs-based OOD Data Classification}
\label{sec:llm_based_ood}
For samples flagged as OOD, TAO-Net harnesses the generative capabilities of LLMs. Specifically, instead of mapping an OOD sample \(X_i\) to a predefined label, we treat it as a text-generation task:
\begin{equation}
P(\hat{y} \mid X_i) \;=\; \prod_{t=1}^T P\bigl(\hat{y}_t \mid X_i, \hat{y}_{<t}\bigr),
\end{equation}
where \(\hat{y}_t\) is the \(t\)-th token of the generated label and \(\hat{y}_{<t}\) are previously generated tokens. By allowing the LLMs to output textual labels, we enable flexible recognition of unknown or newly emerging applications without re-training on novel categories.

\subsection{Semantic-enhanced Prompt Strategy (SPS)}
\label{sec:sps}
To guide the LLM in producing reliable and contextually relevant labels for OOD traffic, we design a Semantic-enhanced Prompt Strategy (SPS). SPS comprises a three-level template structure that balances classification precision and generalization capability:

\begin{itemize}
    \item \textbf{Strict Mode} (\(T_1\)): \(T_1\) only constrains the OOD generation space to specific application categories, aiming to maximize classification precision for a limited set of candidate labels. Formally, \(T_1 = \{\,(X_i, y_i)\;|\; y_i \in Y_{OOD}\}\).
    \item \textbf{Complete Mode} (\(T_2\)): \(T_2\) includes both ID and OOD categories from the current dataset, ensuring a more comprehensive domain knowledge. Here, \(T_2 = \{\,(X_i, y_i)\;|\; y_i \in (Y_{ID} \cup Y_{OOD})\}\).
    \item \textbf{Extended Mode} (\(T_3\)): \(T_3\) further integrates knowledge from multiple datasets, enabling cross-domain generalization. Specifically, \(T_3 = \{\,(X_i, y_i)\;|\; y_i \in \bigcup_{j=1}^n (Y_{ID}^j \cup Y_{OOD}^j)\}\).
\end{itemize}

These modes progressively expand the LLM’s generation space. For a given OOD sample \(X_i\), we construct a prompt \(p = \{T_k, X_i\}\), where \(T_k \in \{T_1, T_2, T_3\}\) is chosen based on the operational scenario. The LLM then produces a textual label \(\hat{y}\) following the conditional distribution \(P(\hat{y}\mid p)\). Experimental evaluations in Section~\ref{sec:sps_analysis} demonstrate that SPS effectively refines the LLM’s outputs and achieves fine-grained classification of previously unseen network traffic.

\medskip

Overall, TAO-Net synergizes a hybrid OOD detection mechanism, a specialized ID classifier, and an LLM-based OOD label generator with SPS. As a result, our framework not only preserves strong performance on known classes but also adaptively recognizes emergent traffic patterns crucial for modern network security applications.

\section{Experiments} \label{sec:experiments}
\subsection{Datasets}
To validate the effectiveness of TAO-Net, experiments are conducted on three public datasets, namely \textbf{CHNAPP}~\cite{peng2024efficiently}, \textbf{ISCXVPN}~\cite{gil2016characterization}, and \textbf{ISCXTor}~\cite{lashkari2017characterization}. These datasets cover diverse real-world encryption scenarios, including common Chinese Internet applications (CHNAPP), VPN-encrypted traffic (ISCXVPN), and Tor network traffic (ISCXTor). 

\noindent\textbf{Preprocessing and ID/OOD Partition.} In order to design suitable ID/OOD evaluation settings, each dataset undergoes a series of preprocessing steps (e.g., selecting categories, packet collecting, etc.). Specifically:
\begin{itemize}
    \item \textbf{CHNAPP} includes traffic from six Chinese Internet applications. Four categories (\emph{QQMail, QQMusic, Youku, TaoBao}) are selected as ID, and two categories (\emph{WeChat, Weibo}) serve as OOD. 
    \item \textbf{ISCXVPN} provides traffic from 13 applications under VPN encryption. Nine categories (\emph{Gmail, Facebook, FTPS, Hangouts, Hangout, Netflix, BitTorrent, SFTP, Skype}) are designated as ID, while four (\emph{VoipBuster, YouTube, Vimeo, Spotify}) are OOD.
    \item \textbf{ISCXTor} contains Tor-encrypted traffic from 12 applications. Eight categories (\emph{Gmail, Facebook, FTP, Hangout, P2P, POP, Skype, Spotify}) are chosen as ID, and four (\emph{SSL, Thunderbird, Vimeo, YouTube}) as OOD.
\end{itemize}

Table~\ref{tab:dataset_statistics} presents the overall data volume and split details for each dataset. Note that only ID samples are included in the training set. Validation and test sets both contain ID and OOD samples with a 7:3 ratio, ensuring robust evaluation of the proposed method under OOD conditions. Specifically, CHNAPP is partitioned into 4 ID classes and 2 OOD classes, ISCXVPN into 9/4, and ISCXTor into 8/4. This setup simulates realistic deployment scenarios, where emergent applications (OOD) appear alongside known ones (ID).

\subsection{Evaluation Metrics}
The evaluation employs four metrics: Macro Precision (M-Prec), Macro F1, Micro F1, and Recall. M-Prec and Macro F1 give equal weight to each class regardless of sample size, while Micro F1 provides a frequency-weighted measure. Recall reflects the ability to identify all relevant instances, crucial for network security applications. Confusion matrices provide detailed performance analysis.

\subsection{Experimental Setting}
\textbf{Training Configuration.} The OOD detection stage uses Adam optimizer (lr=2e-5) for 20 epochs with BCEWithLogitsLoss. The hybrid score parameters are set to $\alpha$=0.6 and $\delta$=0.75. For classification, the ID branch uses AdamW optimizer (lr=2e-5) for 30 epochs, while the OOD branch employs GPT-4o with temperature=0.7 and top-p=0.95.

\begin{table}[t]
\centering
\caption{Dataset statistics for CHNAPP,ISCXVPN,ISCXTor.}
\label{tab:dataset_statistics}
\begin{tabular}{lrrrr}
\toprule
\textbf{Dataset} & \textbf{Total} & \textbf{Train} & \textbf{Valid} & \textbf{Test} \\
\midrule
CHNAPP   & 614,575   & 485,782   & 64,391  & 64,392 \\
ISCXVPN  & 492,598   & 443,337   & 24,631  & 24,630 \\
ISCXTor  & 1,287,303 & 450,001   & 82,287  & 82,287 \\
\bottomrule
\end{tabular}
\end{table}

\noindent\textbf{Implementation Details.} All experiments run on an NVIDIA RTX 4090 GPU with random seed 42, reporting means and standard deviations over 5 independent runs.

\begin{table*}[t]
  \centering
  \caption{Performance comparison of TAO-Net against ID traffic classification models, general language models, and LLM‑based approaches on encrypted traffic datasets. The evaluation metrics include Macro Precision (M‑Prec), Macro F1, Micro F1, and Recall. Bold values indicate the best performance.}
  \label{tab:comparison_results}
  \makebox[\textwidth][c]{%
    \begin{adjustbox}{max width=\textwidth} 
      \begin{tabular}{lcccccccccccc}
        \toprule
        \textbf{Model}
        & \multicolumn{4}{c}{\textbf{CHNAPP}}
        & \multicolumn{4}{c}{\textbf{ISCXVPN}}
        & \multicolumn{4}{c}{\textbf{ISCXTor}}                                                \\
        \cmidrule(lr){2-5} \cmidrule(lr){6-9} \cmidrule(lr){10-13}
        & \textbf{M‑Prec} & \textbf{Macro F1} & \textbf{Micro F1} & \textbf{Recall}
        & \textbf{M‑Prec} & \textbf{Macro F1} & \textbf{Micro F1} & \textbf{Recall}
        & \textbf{M‑Prec} & \textbf{Macro F1} & \textbf{Micro F1} & \textbf{Recall} \\ \toprule
        PacRep      & 0.5813 & 0.5688 & 0.6925 & 0.5573 & 0.5867 & 0.5749 & 0.7003 & 0.5635 & 0.5964 & 0.5841 & 0.7035 & 0.5726 \\
        ET‑BERT     & 0.5503 & 0.5412 & 0.6581 & 0.5325 & 0.6173 & 0.6047 & 0.6864 & 0.5932 & 0.5962 & 0.5829 & 0.6651 & 0.5714 \\ \midrule
        BERT        & 0.5724 & 0.5703 & 0.6992 & 0.5682 & 0.5732 & 0.5624 & 0.6994 & 0.5518 & 0.5763 & 0.5643 & 0.6998 & 0.5532 \\
        BART        & 0.5723 & 0.5695 & 0.6987 & 0.5664 & 0.6207 & 0.6078 & 0.6995 & 0.5964 & 0.5624 & 0.5619 & 0.6998 & 0.5625 \\
        BART‑Large  & 0.5734 & 0.5700 & 0.6987 & 0.5671 & 0.5923 & 0.5803 & 0.6996 & 0.5692 & 0.5864 & 0.5741 & 0.6997 & 0.5625 \\
        T5          & 0.5713 & 0.5687 & 0.6983 & 0.5659 & 0.5704 & 0.5606 & 0.6994 & 0.5527 & 0.5573 & 0.5536 & 0.6994 & 0.5508 \\
        T5‑Large    & 0.5724 & 0.5694 & 0.6985 & 0.5668 & 0.5834 & 0.5715 & 0.6995 & 0.5608 & 0.5803 & 0.5681 & 0.6996 & 0.5573 \\ \midrule
        ChatGLM‑3   & 0.6412 & 0.6280 & 0.6626 & 0.6156 & 0.4473 & 0.4364 & 0.4444 & 0.4256 & 0.5532 & 0.5420 & 0.5764 & 0.5312 \\
        LLaMA‑3     & 0.7283 & 0.7152 & 0.7338 & 0.7035 & 0.5734 & 0.5637 & 0.5717 & 0.5552 & 0.6624 & 0.6520 & 0.6825 & 0.6401 \\
        GPT‑4o      & 0.8630 & 0.8631 & 0.8634 & 0.8637 & 0.8193 & 0.8376 & 0.8326 & 0.8651 & 0.8394 & 0.8513 & 0.8619 & 0.8635 \\ \midrule
        \textbf{TAO‑Net (Ours)}
         & \textbf{0.9681} & \textbf{0.9677} & \textbf{0.9662} & \textbf{0.9678}
         & \textbf{0.9690} & \textbf{0.9683} & \textbf{0.9491} & \textbf{0.9685}
         & \textbf{0.9770} & \textbf{0.9768} & \textbf{0.9597} & \textbf{0.9766} \\ \bottomrule
      \end{tabular}
    \end{adjustbox}
  } 
\end{table*}

\subsection{Baselines}
We compare TAO-Net with nine baselines in three categories:

\textbf{1) ID Traffic Classification:} 
PacRep \cite{meng2022packet} (SOTA with BERT encoder for packet-level representation) and ET-BERT \cite{lin2022bert} (traffic-optimized approach with two-stage training and pre-training on large-scale traffic data).

\textbf{2) General Language Models (GLM):} 
BERT \cite{devlin2018bert} (fundamental transformer model), BART/BART-Large \cite{lewis2019bart} (sequence-to-sequence models), and T5/T5-Large \cite{raffel2020exploring} (text-to-text transfer models).

\textbf{3) LLM-based Approaches:} 
ChatGLM-3 \cite{glm2024chatglm} and LLaMA-3 \cite{touvron2023llama} (with LoRA fine-tuning), and GPT-4o \cite{achiam2023gpt} (zero-shot setting without task-specific fine-tuning).

\subsection{Main Results}

\begin{figure*}
\centering
\includegraphics[width=\textwidth]{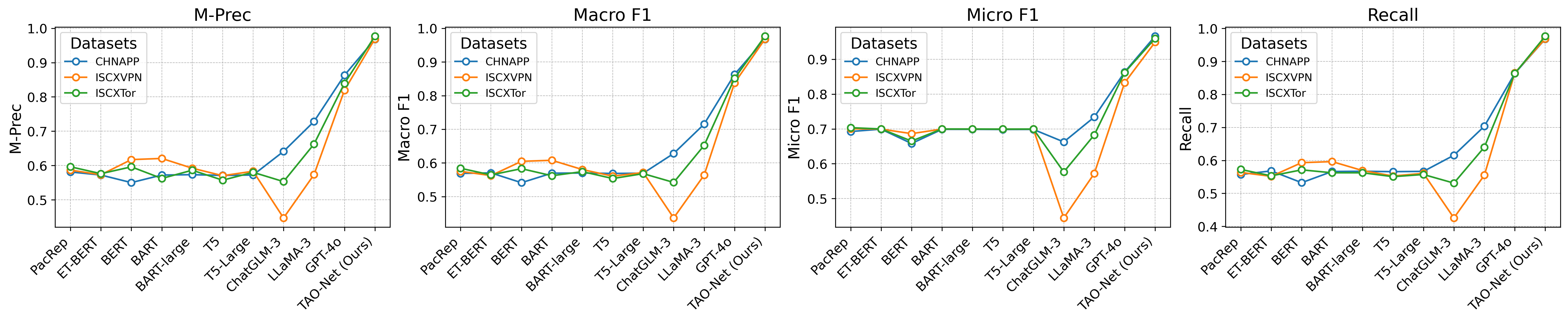}
\caption{Performance comparison of different models across four metrics (M-Prec, Macro F1, Micro F1, and Recall) on three datasets. TAO-Net consistently achieves superior results across all metrics, particularly in handling OOD traffic. The performance gap is most pronounced when classifying VPN-encrypted traffic (ISCXVPN dataset).}
\label{fig:performance_metrics}
\end{figure*}

Table \ref{tab:comparison_results} and Figure \ref{fig:performance_metrics} provide a comprehensive overview of the performance achieved by TAO-Net and various baseline models across three diverse datasets (CHNAPP, ISCXVPN, and ISCXTor). These results underscore several noteworthy observations.

\textbf{1) Consistent Superior Performance Across Metrics.}
As shown in Figure \ref{fig:performance_metrics}, TAO-Net demonstrates consistently superior performance across all evaluation metrics. On the CHNAPP dataset, TAO-Net achieves 96.81\% Macro Precision, significantly surpassing GPT-4o (86.30\%) by over 10 percentage points. Similar substantial advantages are observed in Macro F1 (96.77\% vs 86.31\%) and Micro F1 (96.62\% vs 86.34\%). The consistently high Recall rates (above 96\% across all datasets) clearly highlight TAO-Net's exceptional ability to identify diverse traffic patterns.

\textbf{2) Robust Performance in Complex OOD Scenarios.}
TAO-Net's performance advantage is particularly evident in challenging scenarios, especially with VPN-encrypted traffic (ISCXVPN). While GPT-4o achieves 81.93\% Macro Precision, TAO-Net significantly improves this to 96.90\%. Notable performance gaps are also observed in Macro F1 (96.83\% vs 83.76\%) and Micro F1 (94.91\% vs 83.26\%). These results demonstrate TAO-Net's robust generalization capability when dealing with heavily obfuscated traffic patterns.

\textbf{3) Clear Performance Stratification Among Baselines.}
The results reveal a distinct performance hierarchy among baseline approaches. ID traffic classification models (PacRep and ET-BERT) achieve Macro Precision scores of 55.03-61.73\% across datasets. General language models (BERT, BART, and T5 variants) show minimal improvements, maintaining Macro Precision in the 55.73-62.07\% range. LLM-based approaches demonstrate more substantial capabilities, with GPT-4o leading the baselines (Macro Precision: 81.93-86.30\%). However, even these advanced models fall notably short of TAO-Net's performance, which consistently exceeds 96\% across all datasets.

\textbf{4) Dataset-Specific Performance Variations.}
While TAO-Net maintains exceptional performance across all datasets, there are notable variations in the improvement margins. The model achieves its highest Macro Precision (97.70\%) on the ISCXTor dataset, followed by ISCXVPN (96.90\%) and CHNAPP (96.81\%). This consistent performance above 96\% demonstrates TAO-Net's robust adaptability across different traffic encryption scenarios. Interestingly, the performance gap between TAO-Net and baselines is most pronounced in the ISCXVPN dataset, where TAO-Net outperforms GPT-4o by 14.97 percentage points (96.90\% vs 81.93\%), suggesting particular effectiveness in handling VPN-encrypted traffic.

\textbf{5) Advantages of Generative Classification Approach.}
A striking observation from both Table \ref{tab:comparison_results} and Figure \ref{fig:performance_metrics} is the substantial performance improvement (38.68-41.06 percentage points in Macro Precision) that TAO-Net achieves over previous classification approaches. This significant enhancement can be attributed to our innovative transformation of traffic classification into a generation task within a carefully designed two-stage adaptive framework. Unlike conventional methods that focus on incremental improvements in model capacity, TAO-Net fundamentally reimagines the entire classification process, enabling superior handling of both known and emerging traffic patterns.

The comprehensive quantitative results validate TAO-Net's exceptional capabilities in identifying both ID and OOD traffic under various encryption scenarios. By consistently achieving macro-precision of 96.81-97.70\%, macro-F1 of 96.77-97.68\%, and micro-F1 of 94.91-96.62\% across all evaluation measures, particularly in challenging VPN and Tor networks, TAO-Net demonstrates strong potential for practical deployment in contemporary network environments where encryption complexity and unknown traffic patterns are increasingly prevalent.

\section{Analyses}
\subsection{Ablation Study}

To comprehensively evaluate the effectiveness of each core component in TAO-Net, we conduct systematic ablation studies. These experiments primarily investigate the contribution of the two-stage adaptive design and its advantages in handling OOD traffic.

\paragraph{Experimental Design.}
We design two controlled comparative experiments to isolate and quantify role of each stage:
\begin{itemize}
\item \textbf{TAO-Net with only PacRep:} Retains OOD detector but uses PacRep for all traffic classification, evaluating benefits of OOD detection on previous approaches.
\item \textbf{TAO-Net with only GPT-4o:} Keeps OOD detector but uses GPT-4o exclusively, with ID/OOD information injected via SPS framework.
\end{itemize}

\paragraph{Results and Analysis.} 
Table~\ref{tab:taonet_comparison} presents the ablation results on three datasets (CHNAPP, ISCXVPN, and ISCXTor). We summarize the key findings below.

\textbf{1) Positive Gains from OOD Detection.}  
Comparing \emph{PacRep (Baseline)} with \emph{TAO-Net with only PacRep} shows that introducing OOD detection yields moderate yet consistent improvements. For instance, Macro Precision rises from 58.13\% to 59.07\% on CHNAPP, and from 58.67\% to 59.73\% on ISCXVPN. Although the performance gains may appear incremental, the improvement underscores the benefit of filtering out unknown traffic. Without OOD detection, PacRep must classify unseen OOD data as if it were part of the known categories, thereby introducing additional confusion. Hence, incorporating OOD detection helps reduce misclassification and paves the way for more robust handling of unknown traffic in subsequent stages.

\textbf{2) Substantial Advantage of LLM Integration.}  
When GPT-4o alone is used for all traffic (i.e., \emph{TAO-Net with only GPT-4o}), we observe significant improvements over the baseline PacRep. On CHNAPP, Macro Precision increases markedly from 58.13\% to 88.24\%, with similar rises in ISCXVPN (85.34\%) and ISCXTor (86.32\%). This boost primarily stems from the strong zero-shot generalization ability inherent in large language models. Moreover, our carefully designed \textbf{SPS} (Semantic-enhanced Prompt Strategy) further enhances label generation by structuring the prompt templates in a way that narrows down the plausible output space for each traffic scenario. In this manner, the LLM leverages its extensive language understanding to classify unknown traffic patterns more accurately.

\textbf{3) Full TAO-Net Framework Achieves Outstanding Performance.}  
The complete TAO-Net setup significantly outperforms the single-model variants across all metrics and datasets. For example, in the CHNAPP dataset, TAO-Net achieves a Macro Precision of 96.81\%, outperforming both the PacRep-only variant (58.13\%) and GPT-4o variant (86.30\%). Similar advantages are evident in ISCXVPN (96.90\% vs 58.67\% and 81.93\%) and ISCXTor (97.70\% vs 59.64\% and 83.94\%). Notably, TAO-Net consistently achieves Macro F1 (96.77-97.68\%), Micro F1 (94.91-96.62\%), and Recall (96.78-97.66\%) values exceeding 94\%, demonstrating strong robustness across different traffic encryption scenarios.

\textbf{4) Synergistic Effect of Components.}  
The ablation results highlight how TAO-Net's two-stage architecture and integrated components together deliver superior results. Concretely, our full model surpasses the baseline PacRep by 38.68-41.06 percentage points in Macro Precision (rising from 58.13-59.64\% to 96.81-97.70\%) and further outperforms GPT-4o variant by 10.51-14.97\%. These substantial improvements attest that accurate OOD detection in first stage, followed by different processing in second stage (PacRep for ID data and LLM for OOD data), is essential for achieving robust encrypted traffic classification. The interplay of both stages reduces misclassification risk, allowing TAO-Net to excel under diverse real-world network conditions.

Overall, this ablation study confirms the critical roles of each component in TAO-Net. By effectively integrating OOD detection, a dedicated ID classifier, and an LLM-based OOD classifier under the SPS framework, TAO-Net meets the demanding requirements of modern network environments where continuous emergence of new traffic types is the norm.

\begin{table*}[t]
\centering
\caption{Ablation Study Results of TAO-Net: Comparative Analysis Across Different Datasets Demonstrating Component-wise Contributions. Evaluation metrics include M-Prec, Macro F1, Micro F1, and Recall. Bold values indicate the best performance.} 
\label{tab:taonet_comparison}
\begin{adjustbox}{max width=\textwidth}
\begin{tabular}{lcccccccccccc}
\toprule
\textbf{Model} 
& \multicolumn{4}{c}{\textbf{CHNAPP}} 
& \multicolumn{4}{c}{\textbf{ISCXVPN}} 
& \multicolumn{4}{c}{\textbf{ISCXTor}} \\
\cmidrule(lr){2-5} \cmidrule(lr){6-9} \cmidrule(lr){10-13}
 & \textbf{M-Prec} & \textbf{Macro F1} & \textbf{Micro F1} & \textbf{Recall}
 & \textbf{M-Prec} & \textbf{Macro F1} & \textbf{Micro F1} & \textbf{Recall}
 & \textbf{M-Prec} & \textbf{Macro F1} & \textbf{Micro F1} & \textbf{Recall} \\
\midrule
PacRep (Baseline)        
& 0.5813 & 0.5688 & 0.6925 & 0.5573   
& 0.5867 & 0.5749 & 0.7003 & 0.5635   
& 0.5964 & 0.5841 & 0.7035 & 0.5726 \\
    
~~ w/ only PacRep   
& 0.5907 & 0.5695 & 0.6938 & 0.5585   
& 0.5973 & 0.5756 & 0.7008 & 0.5642   
& 0.6062 & 0.5848 & 0.7042 & 0.5734 \\
    
~~ w/ only GPT-4o   
& 0.8824 & 0.8778 & 0.8782 & 0.8746   
& 0.8534 & 0.8492 & 0.8483 & 0.8445   
& 0.8632 & 0.8622 & 0.8624 & 0.8586 \\ \midrule
    
\textbf{TAO-Net (Full)}  
& \textbf{0.9681} & \textbf{0.9677} & \textbf{0.9662} & \textbf{0.9678}  
& \textbf{0.9690} & \textbf{0.9683} & \textbf{0.9491} & \textbf{0.9685}  
& \textbf{0.9770} & \textbf{0.9768} & \textbf{0.9597} & \textbf{0.9766} \\
\bottomrule
\end{tabular}
\end{adjustbox}
\end{table*}

\begin{figure*}
\centering
\begin{subfigure}[b]{0.24\textwidth}
    \centering
    \includegraphics[width=\textwidth]{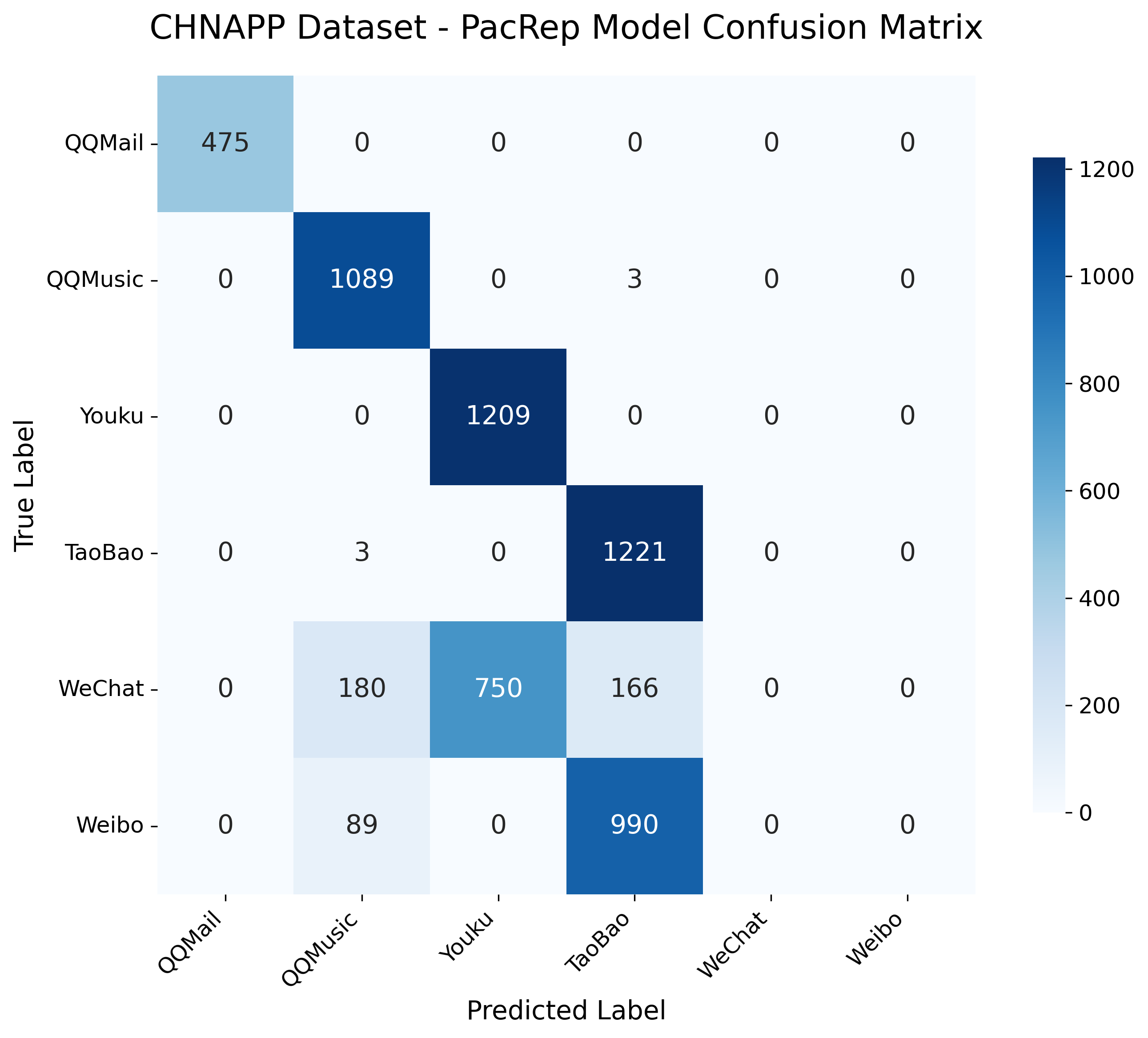}
    \caption{PacRep on CHNAPP}
    \label{fig:chnapp_pacrep}
\end{subfigure}
\begin{subfigure}[b]{0.24\textwidth}
    \centering
    \includegraphics[width=\textwidth]{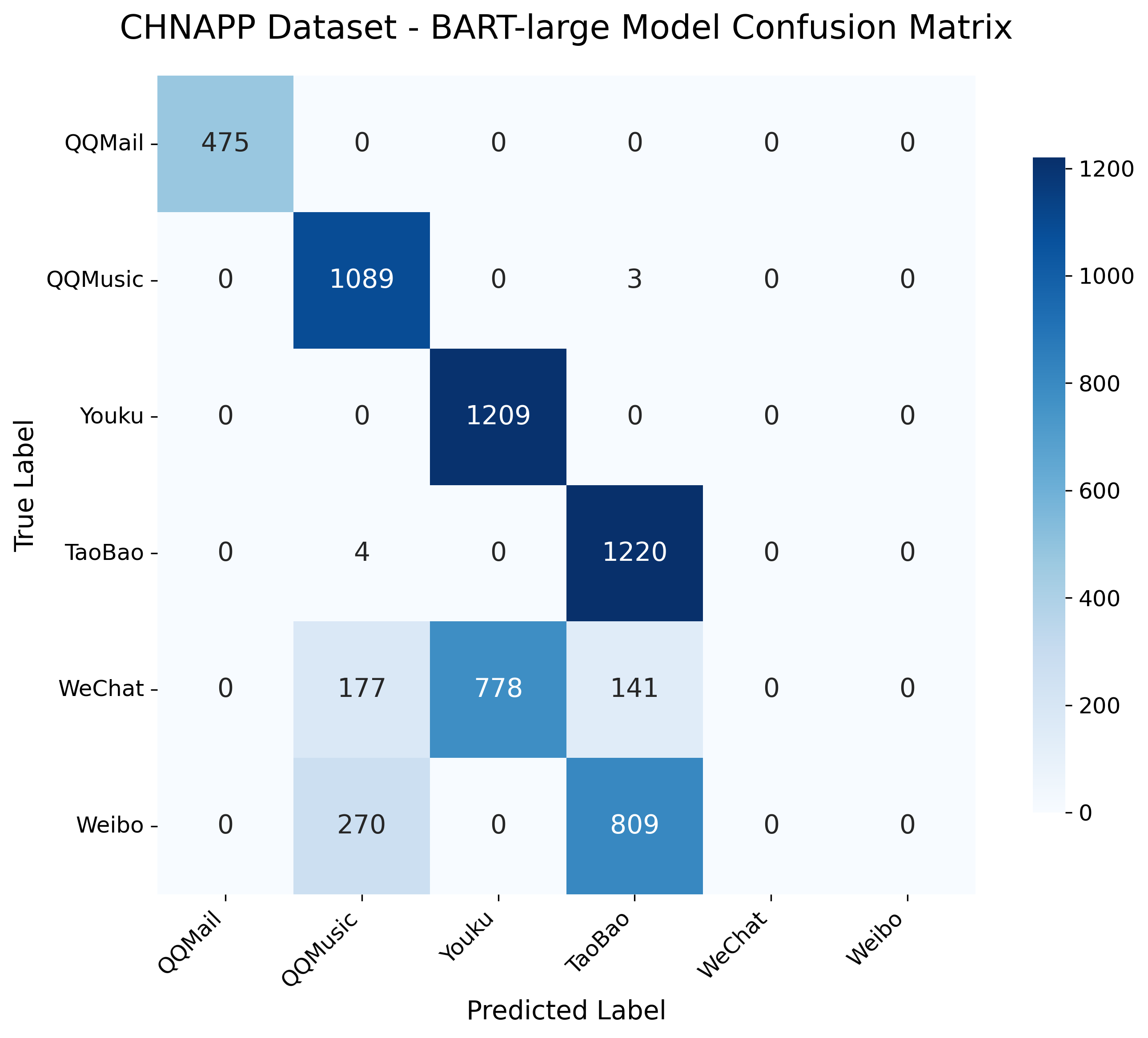}
    \caption{BART-Large on CHNAPP}
    \label{fig:chnapp_bart}
\end{subfigure}
\begin{subfigure}[b]{0.24\textwidth}
    \centering
    \includegraphics[width=\textwidth]{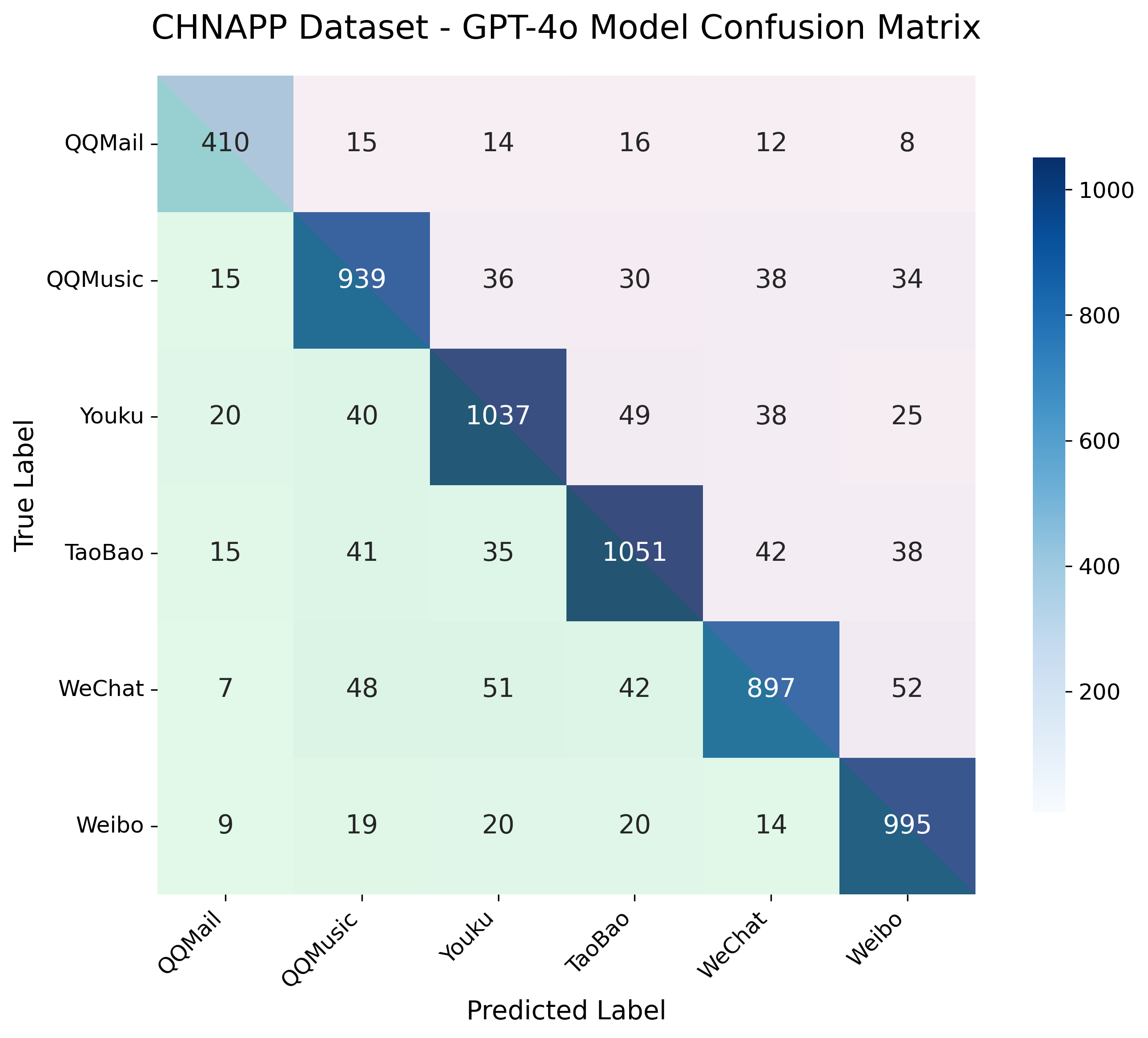}
    \caption{GPT-4o on CHNAPP}
    \label{fig:chnapp_gpt4o}
\end{subfigure}
\begin{subfigure}[b]{0.24\textwidth}
    \centering
    \includegraphics[width=\textwidth]{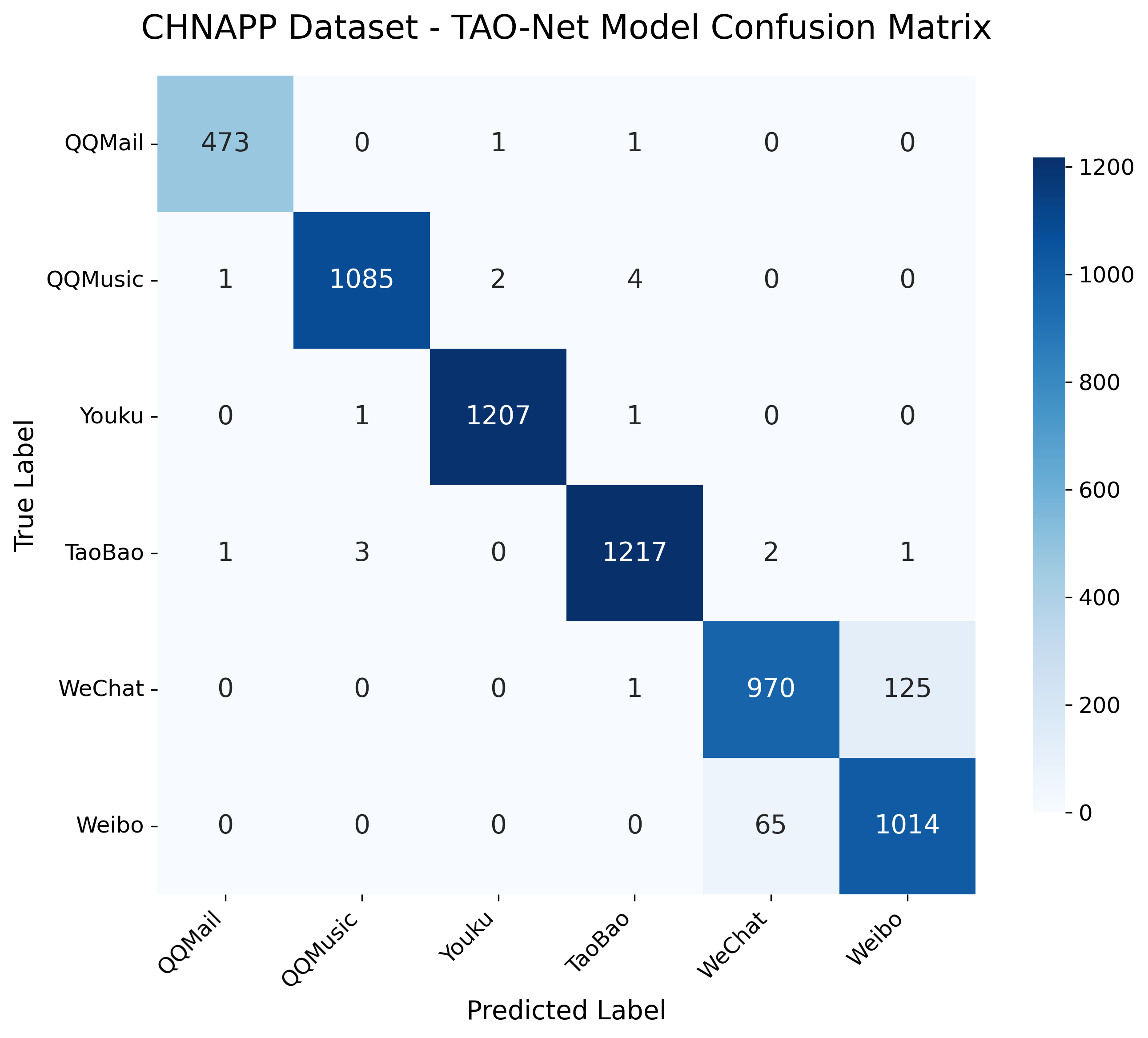}
    \caption{TAO-Net on CHNAPP}
    \label{fig:chnapp_taonet}
\end{subfigure}

\begin{subfigure}[b]{0.24\textwidth}
    \centering
    \includegraphics[width=\textwidth]{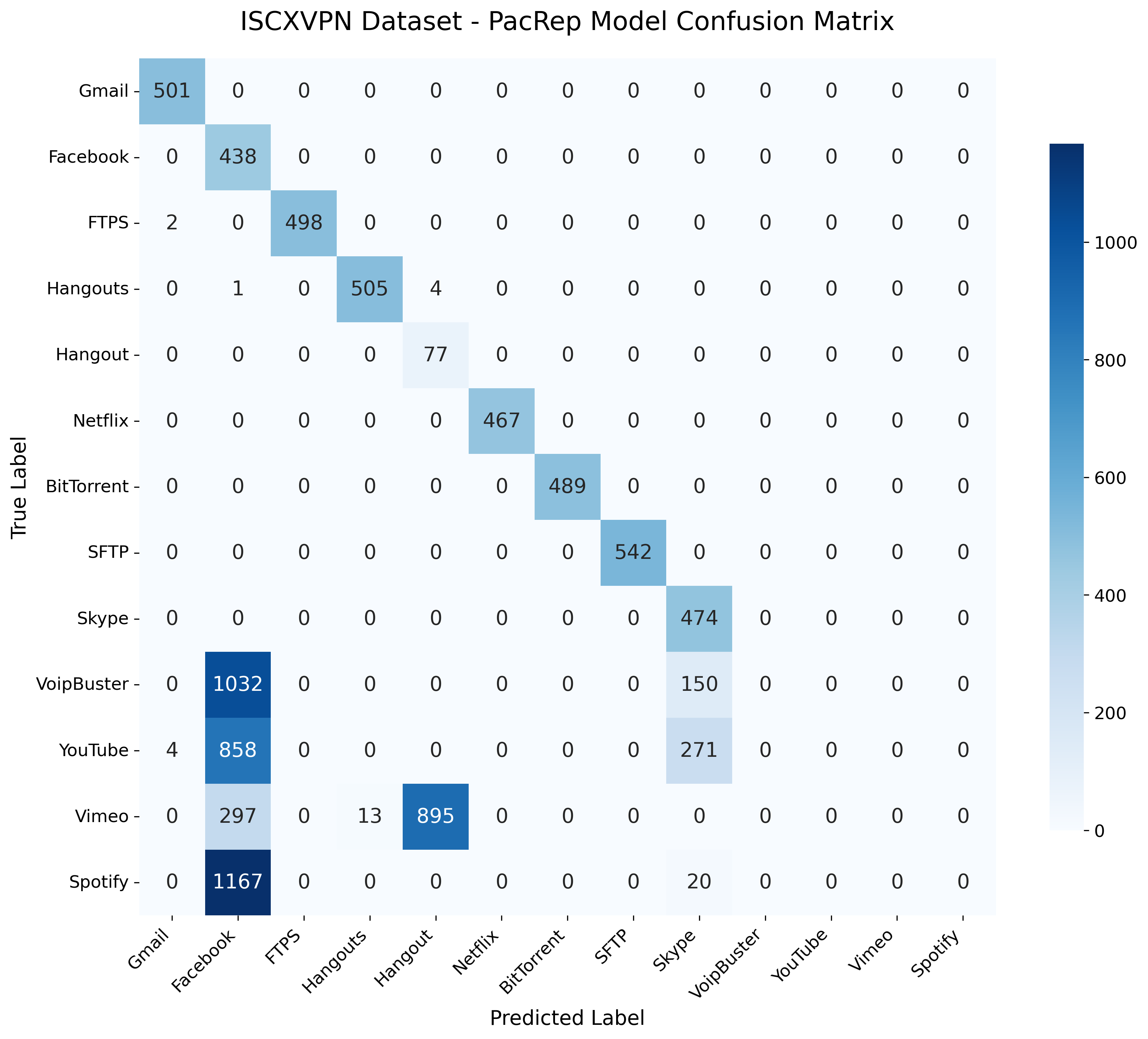}
    \caption{PacRep on ISCXVPN}
    \label{fig:iscxvpn_pacrep}
\end{subfigure}
\begin{subfigure}[b]{0.24\textwidth}
    \centering
    \includegraphics[width=\textwidth]{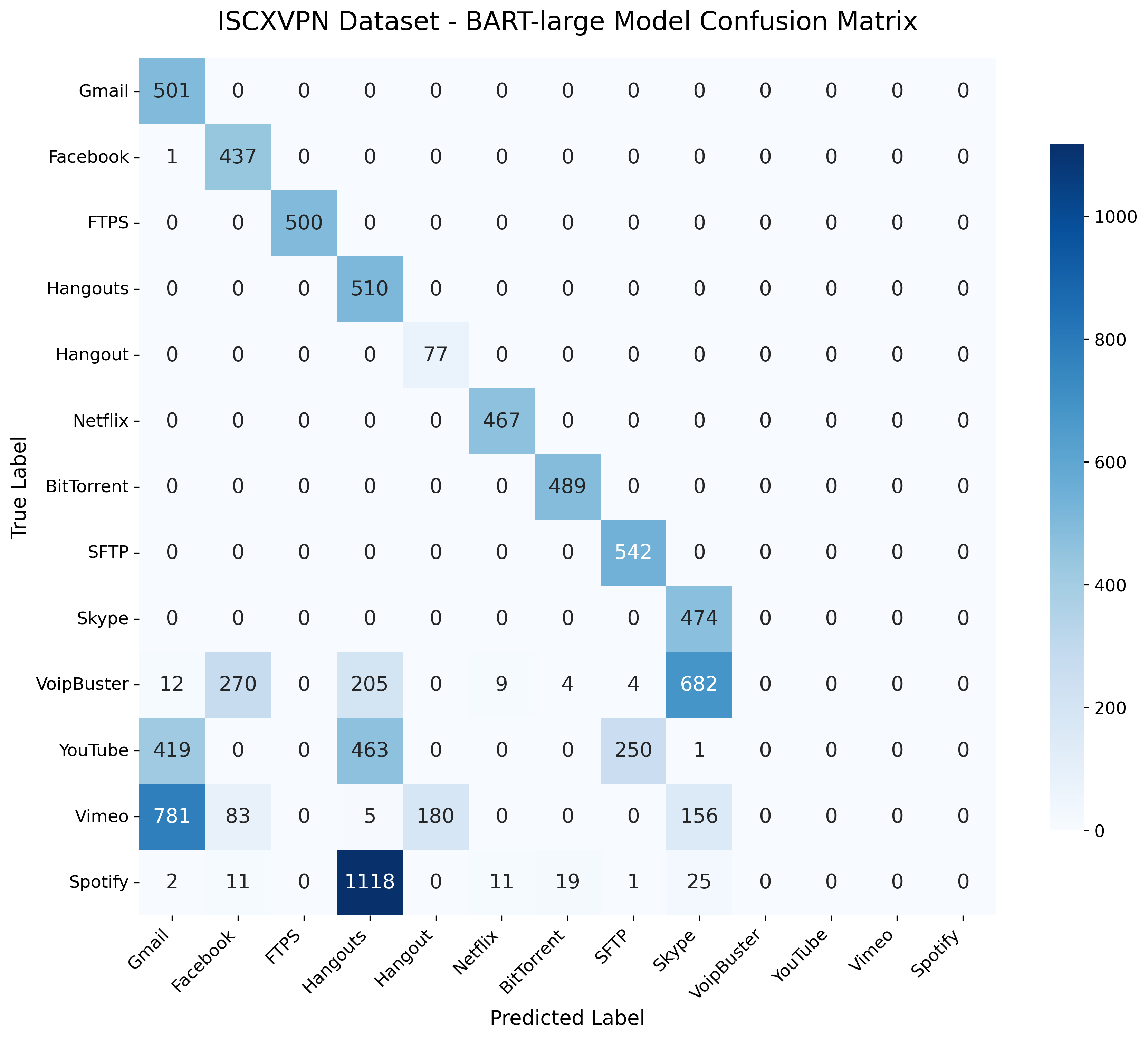}
    \caption{BART-Large on ISCXVPN}
    \label{fig:iscxvpn_bart}
\end{subfigure}
\begin{subfigure}[b]{0.24\textwidth}
    \centering
    \includegraphics[width=\textwidth]{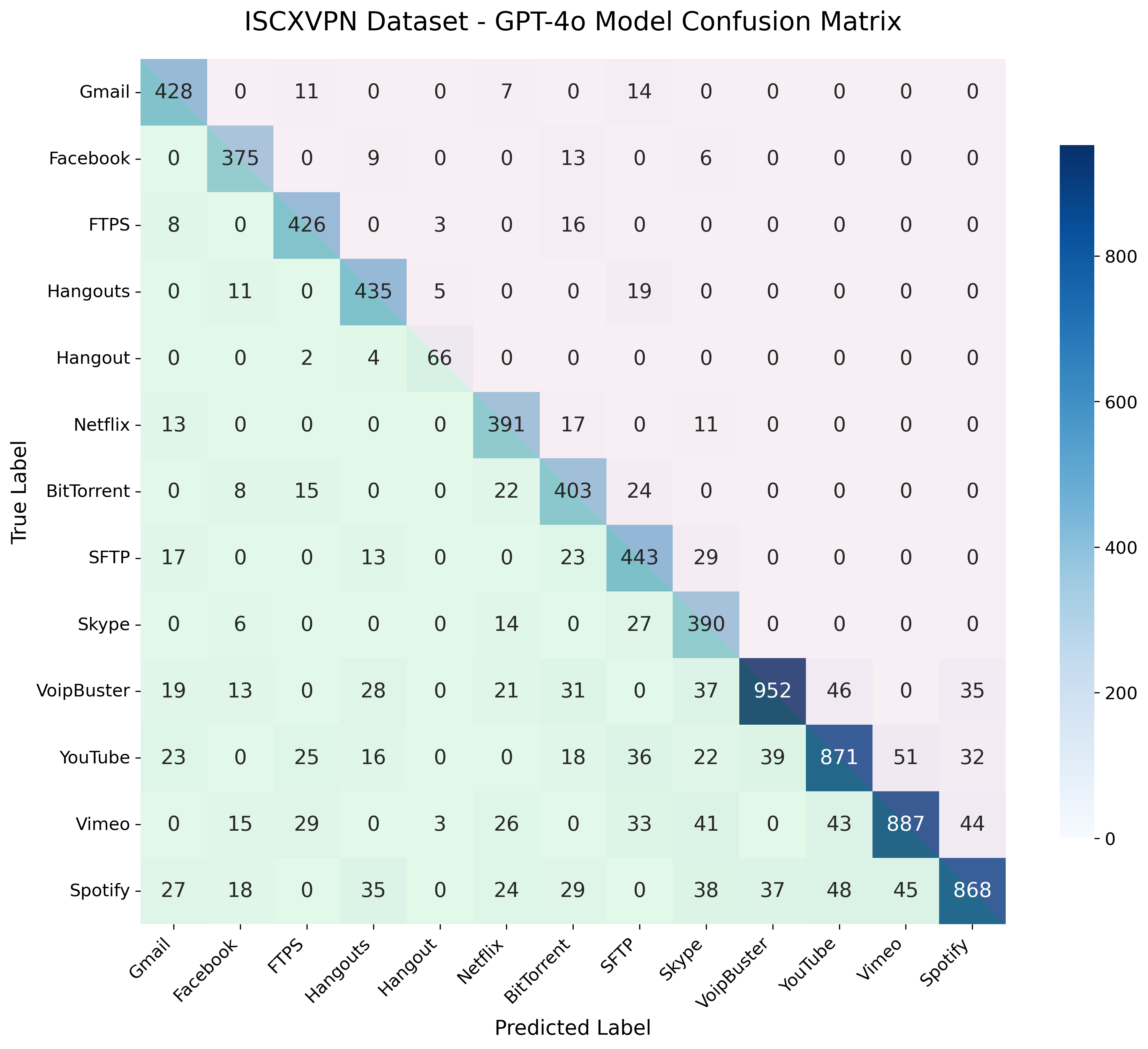}
    \caption{GPT-4o on ISCXVPN}
    \label{fig:iscxvpn_gpt4o}
\end{subfigure}
\begin{subfigure}[b]{0.24\textwidth}
    \centering
    \includegraphics[width=\textwidth]{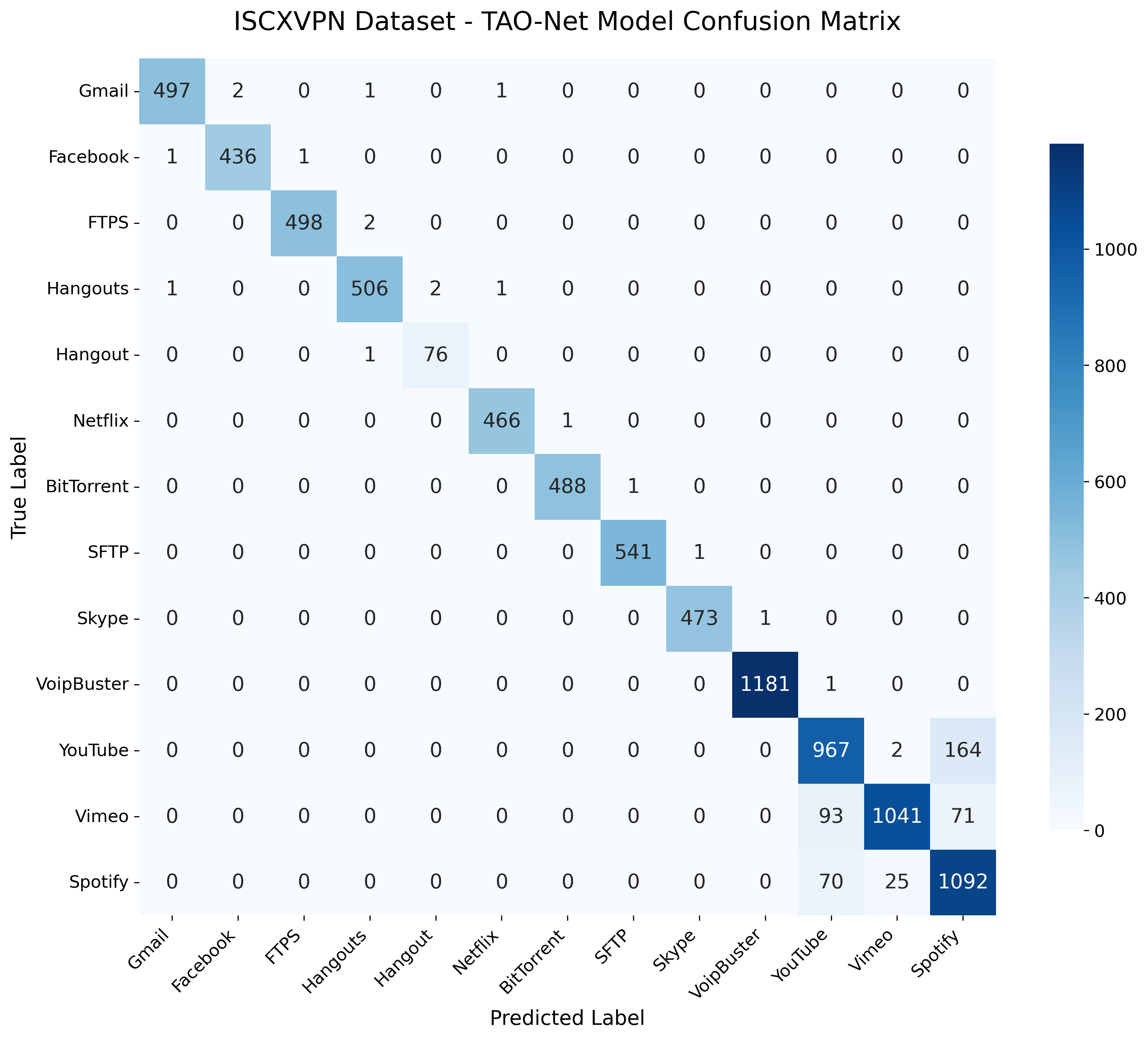}
    \caption{TAO-Net on ISCXVPN}
    \label{fig:iscxvpn_taonet}
\end{subfigure}

\begin{subfigure}[b]{0.24\textwidth}
    \centering
    \includegraphics[width=\textwidth]{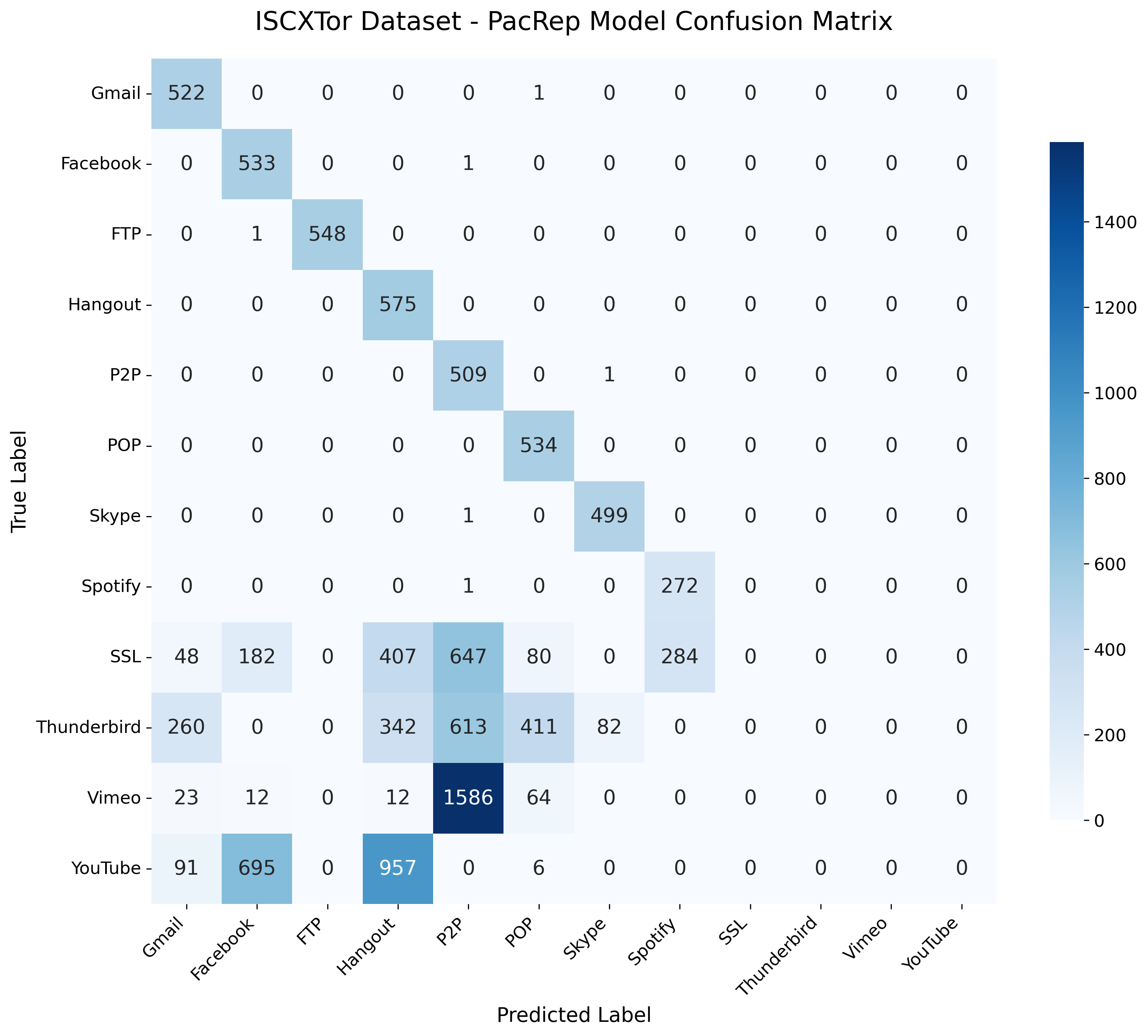}
    \caption{PacRep on ISCXTor}
    \label{fig:iscxtor_pacrep}
\end{subfigure}
\begin{subfigure}[b]{0.24\textwidth}
    \centering
    \includegraphics[width=\textwidth]{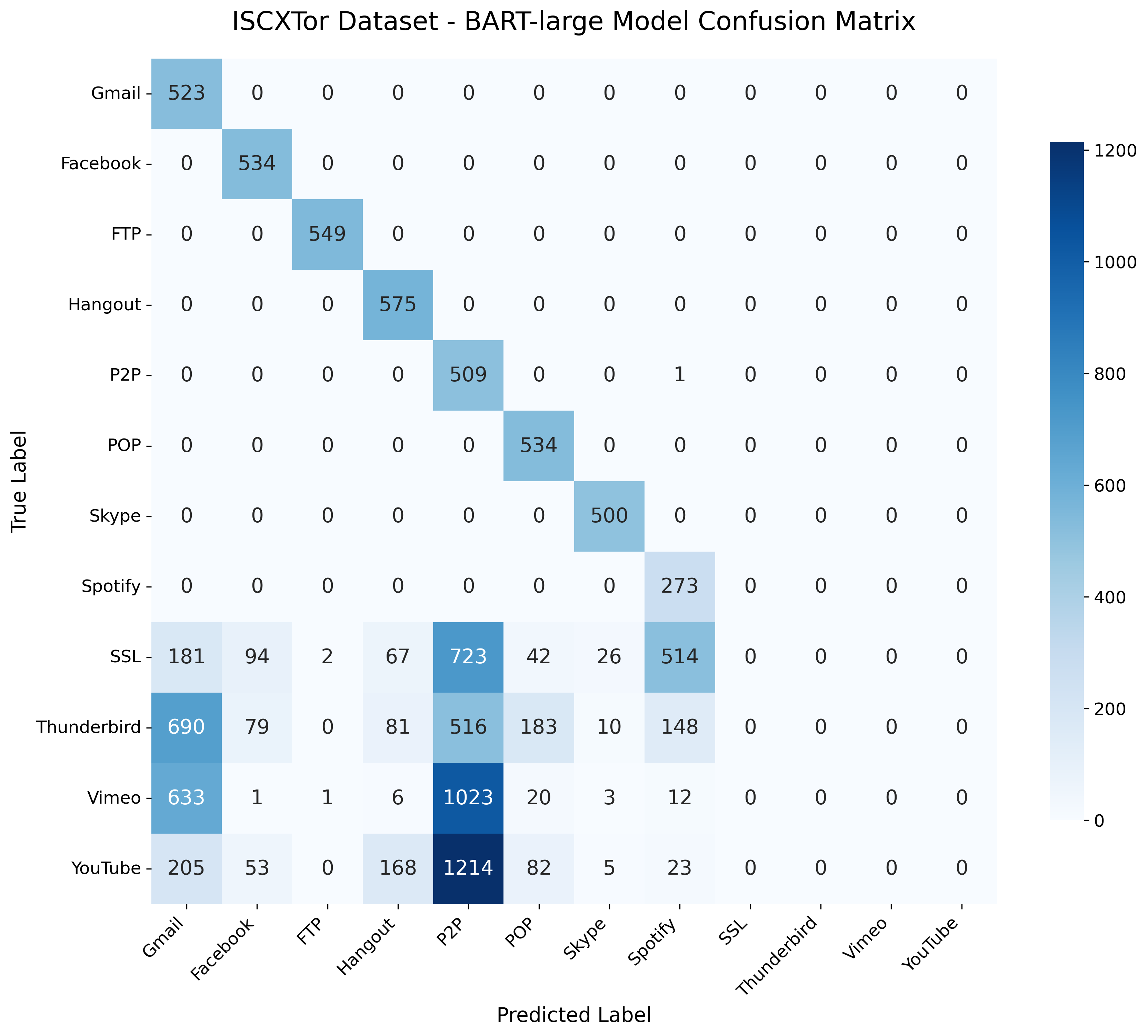}
    \caption{BART-Large on ISCXTor}
    \label{fig:iscxtor_bart}
\end{subfigure}
\begin{subfigure}[b]{0.24\textwidth}
    \centering
    \includegraphics[width=\textwidth]{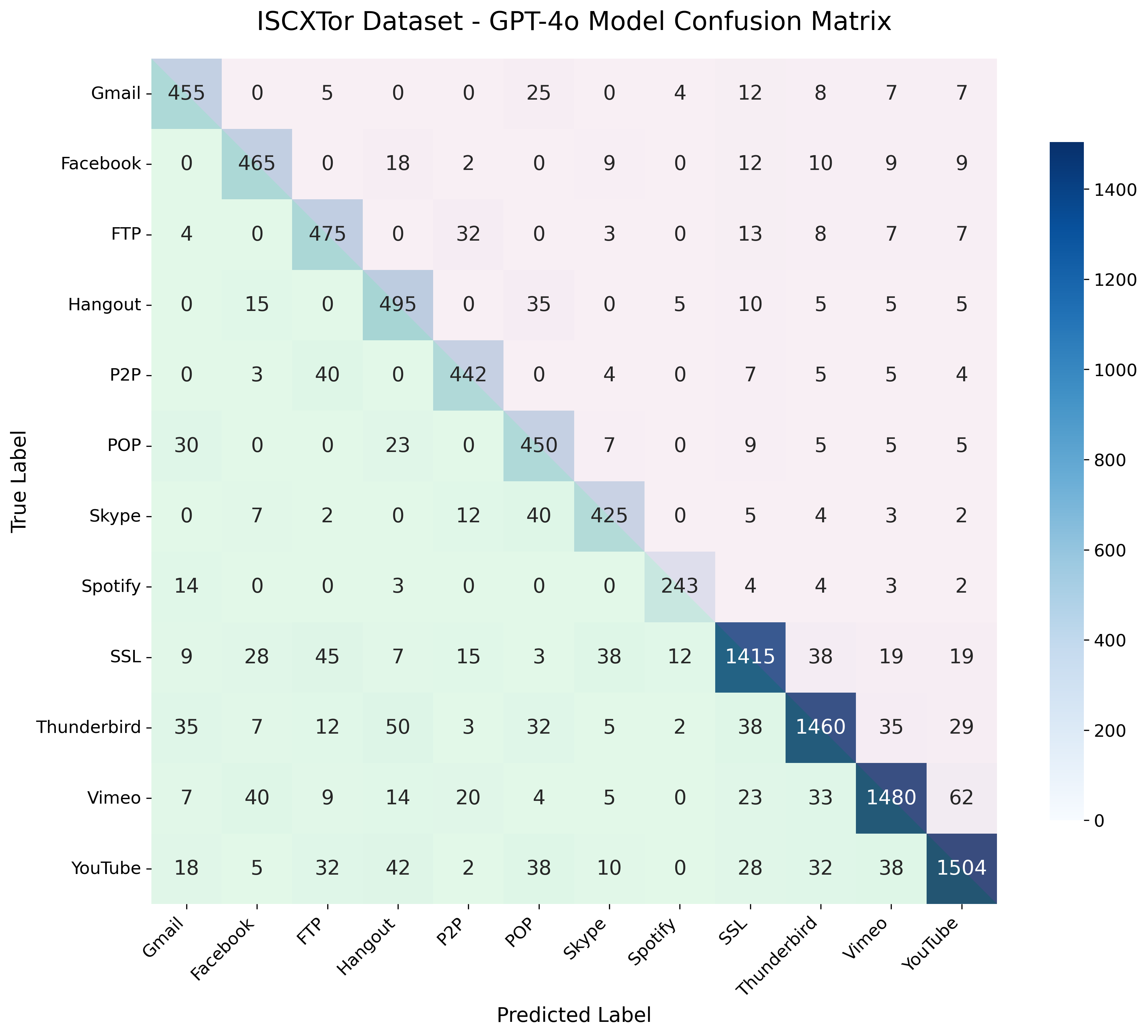}
    \caption{GPT-4o on ISCXTor}
    \label{fig:iscxtor_gpt4o}
\end{subfigure}
\begin{subfigure}[b]{0.24\textwidth}
    \centering
    \includegraphics[width=\textwidth]{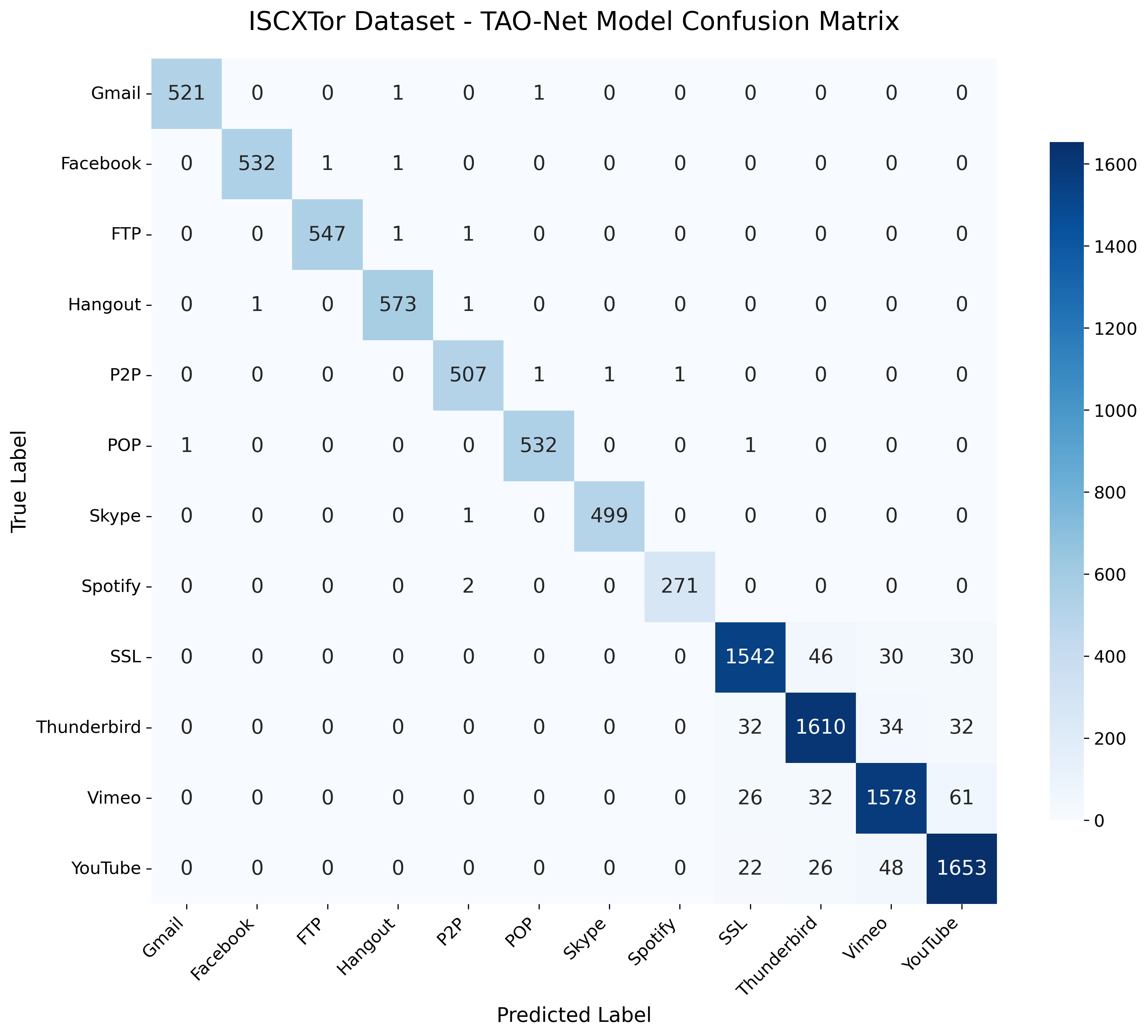}
    \caption{TAO-Net on ISCXTor}
    \label{fig:iscxtor_taonet}
\end{subfigure}

\caption{Confusion matrices comparing model performance across CHNAPP, ISCXVPN, and ISCXTor datasets. Darker colors indicate higher prediction accuracy. The red and green boxes highlight numerous non-zero elements in GPT-4o's upper/lower triangular regions due to lack of OOD detection (Stage 1), while TAO-Net's two-stage design maintains predominantly zero values in these regions.}
\label{fig:confusion_matrices}
\end{figure*}

\subsection{Confusion Matrix Analysis}
To provide a more detailed and intuitive understanding of model performance across different traffic categories, we visualize the confusion matrices for PacRep (best performing ID traffic classification model), BART-Large (best performing GLM-based method), GPT-4o (best LLM-based approach), and our proposed TAO-Net across all three datasets. Figure \ref{fig:confusion_matrices} presents these comparative results.

Analysis of these confusion matrices reveals several critical insights about model performance:

\textbf{1) Limitations of Previous ID Traffic Classification:}  
The confusion matrices for PacRep demonstrate significant limitations in handling OOD traffic. Although PacRep shows relatively stable performance on ID categories (with clear diagonal patterns in the upper-left quadrant), it completely fails to identify OOD traffic categories. For example, in the CHNAPP dataset (Figure \ref{fig:chnapp_pacrep}), PacRep misclassifies a large portion of WeChat traffic (180 instances as QQMusic, 750 as Youku, 166 as TaoBao), indicating its inability to distinguish new application patterns. This failure stems from two key limitations: (1) the model’s rigid category-specific feature extractors, trained only on ID classes, fail to capture OOD traffic; (2) the lack of an OOD detection mechanism forces all traffic into predefined categories, leading to high misclassification rates for unknown applications.

\textbf{2) Inadequate Generalization of GLM-based Methods:}  
BART-Large's confusion matrices reveal severe limitations faced by GLM-based methods. Despite leveraging pre-trained language understanding capabilities, these models completely fail to identify OOD traffic. In the CHNAPP dataset (Figure \ref{fig:chnapp_bart}), all OOD traffic instances are misclassified into ID categories, showing no ability to recognize new traffic patterns. This can be attributed to: (1) limited parameter space (400M parameters) restricting the ability to learn complex traffic characteristics; (2) the lack of domain-specific constraints and OOD detection mechanisms inevitably leading to forced categorization into known classes.

\textbf{3) Improved but Incomplete Performance of Direct LLM Application:} GPT-4o demonstrates notable improvements over previous baselines, particularly in handling OOD traffic. In the CHNAPP dataset (Figure \ref{fig:chnapp_gpt4o}), it achieves better accuracy for most categories (e.g., 939 correct predictions for QQMusic, 1037 correct predictions for Youku). However, several critical limitations persist, especially evident in the visualization patterns shown in Figure \ref{fig:confusion_matrices} (see green and pink part). When examining the upper and lower triangular regions of GPT-4o's confusion matrices, we observe numerous non-zero elements indicating substantial cross-category misclassification, in stark contrast to TAO-Net's clear separation with predominantly zero values in these regions.

These limitations primarily stem from three fundamental aspects of direct LLM application to traffic classification:

a) \textbf{Lack of Domain-Specific Architectural Constraints:} While GPT-4o's transformer architecture excels at capturing general sequential patterns, it lacks built-in constraints for handling network traffic's unique characteristics. The model processes traffic data as generic sequences without incorporating domain-specific priors about protocol structures or network behaviors. This architectural limitation is evident in the confusion matrices, where we observe scattered misclassifications across multiple categories (e.g., VoIP traffic distributed across different labels in Figure \ref{fig:iscxvpn_gpt4o}, red box), indicating the model's inability to maintain consistent classification boundaries based on traffic-specific features.

b) \textbf{Suboptimal Feature Space Mapping:} GPT-4o's pre-training focuses on natural language understanding, creating a feature space that may not optimally capture network traffic patterns. The model maps traffic into this language-oriented space, potentially losing critical protocol-level information. This manifests in the confusion matrices as clusters of misclassifications between similar applications (e.g., streaming services like YouTube and Vimeo, green box in Figure \ref{fig:iscxvpn_gpt4o}), suggesting the model struggles to distinguish fine-grained protocol-specific features.

c) \textbf{Absence of Hierarchical Decision Making:} Unlike TAO-Net's two-stage approach, GPT-4o attempts to directly classify all traffic in a single step. This forces the model to simultaneously handle both the ID/OOD distinction and fine-grained classification, leading to increased uncertainty and error propagation. The impact is visible in the confusion matrices through the presence of significant off-diagonal elements across all categories, indicating the model's difficulty in maintaining clear decision boundaries.

In contrast, TAO-Net addresses these limitations through:

a) \textbf{Specialized Processing Pathways:} By separating ID and OOD detection into distinct stages, TAO-Net can apply targeted processing strategies optimized for each traffic type. This architectural design eliminates the feature space confusion seen in GPT-4o, as evidenced by the clear, block-diagonal patterns in TAO-Net's confusion matrices.

b) \textbf{Enhanced Feature Utilization:} TAO-Net's hybrid OOD detection mechanism focuses on traffic-relevant features like inter-layer transformation smoothness and protocol characteristics. This domain-aware feature processing leads to more robust classification boundaries, visible in the minimal off-diagonal elements in TAO-Net's confusion matrices.

c) \textbf{Guided Generation Process:} For OOD traffic, TAO-Net's Semantic-enhanced Prompt Strategy (SPS) provides structured guidance for the generation process, effectively constraining the output space based on network domain knowledge. This results in more precise classification of unknown applications, as shown by the clear separation between ID and OOD categories in the confusion matrices.

\begin{table*}[t]
\centering
\caption{Comparison of Different SPS Modes for Network Traffic Classification. Strict mode constrains the generation space to specific application categories, Complete mode includes all dataset-specific applications, while Extended mode incorporates cross-dataset knowledge.}
\label{tab:prompt_comparison_llm}
\renewcommand{\arraystretch}{1.14}
\begin{tabular}{l p{15cm}}
\toprule
\textbf{Strategy} & \textbf{Prompt Template} \\
\midrule
\textbf{Strict} & \textit{``Classify this encrypted network traffic packet into one of these known application categories: QQMail, QQMusic, Youku, TaoBao. Consider the traffic packet characteristics including: 1. Protocol behavior (TCP flags, window size) 2. Packet structure (length, fragmentation) 3. Encrypted payload patterns. Your output should be exactly one application name without any additional explanation.''} \\
\midrule
\textbf{Complete} & \textit{``Given this encrypted network traffic packet, classify it into one of these possible applications: QQMail, QQMusic, Youku, TaoBao, WeChat, Weibo. Analyze the characteristics including TCP protocol behaviors, packet structure patterns, and encrypted payload features. Consider both transport layer behaviors and application layer patterns when making prediction. Your output should be exactly one application name.''} \\
\midrule
\textbf{Extended} & \textit{``Analyze this encrypted network traffic packet and classify it based on its characteristics. Consider the following applications across different platforms: QQMail, QQMusic, Youku, TaoBao, WeChat, Weibo, Gmail, Facebook, Skype, YouTube. Examine the traffic packet features including protocol behaviors (e.g., TCP flags, window size), packet structures (length, fragmentation), and encrypted payload patterns. Provide exactly one application name as output.''} \\
\bottomrule
\end{tabular}
\end{table*}

\textbf{4) Comprehensive Excellence of TAO-Net:} 
TAO-Net demonstrates superior performance through its innovative two-stage design, as clearly evidenced by several key patterns. The confusion matrices show distinct, high-accuracy blocks for both ID and OOD traffic types. In CHNAPP dataset (Figure \ref{fig:chnapp_taonet}), we observe exceptional accuracy for ID categories (1085 correct predictions for QQMusic, 1207 correct predictions for Youku) and OOD categories (970 for WeChat, 1014 for Weibo).

The synergistic effect of the two-stage design is particularly visible in the confusion matrices' off-diagonal regions. Unlike GPT-4o, which shows scattered misclassifications between ID and OOD categories, TAO-Net maintains clear boundaries between these categories with minimal cross-category confusion. This improvement stems from the framework's ability to first identify the traffic type (ID/OOD) before applying specialized processing, effectively reducing the search space and improving classification precision.

In the challenging ISCXVPN dataset (Figure \ref{fig:iscxvpn_taonet}), TAO-Net's two-stage architecture demonstrates its strength in handling complex encrypted traffic. The confusion matrix shows remarkably clean diagonal patterns across both ID and OOD categories, with minimal off-diagonal noise. This is particularly noteworthy for similar application types where previous models, including GPT-4o, showed significant confusion. The accurate classification of these similar services highlights how the initial ID/OOD separation enables more nuanced feature analysis in the second stage.

The ISCXTor results (Figure \ref{fig:iscxtor_taonet}) further illustrate the architecture's effectiveness in handling multi-layered encryption scenarios. The confusion matrix exhibits clear, dark diagonal elements with notably reduced off-diagonal values compared to all baselines. This performance advantage stems from the framework's ability to leverage different classification strategies for ID and OOD traffic while maintaining consistent feature interpretation across both stages.

\begin{figure*}
\centering
\includegraphics[width=0.999\textwidth]{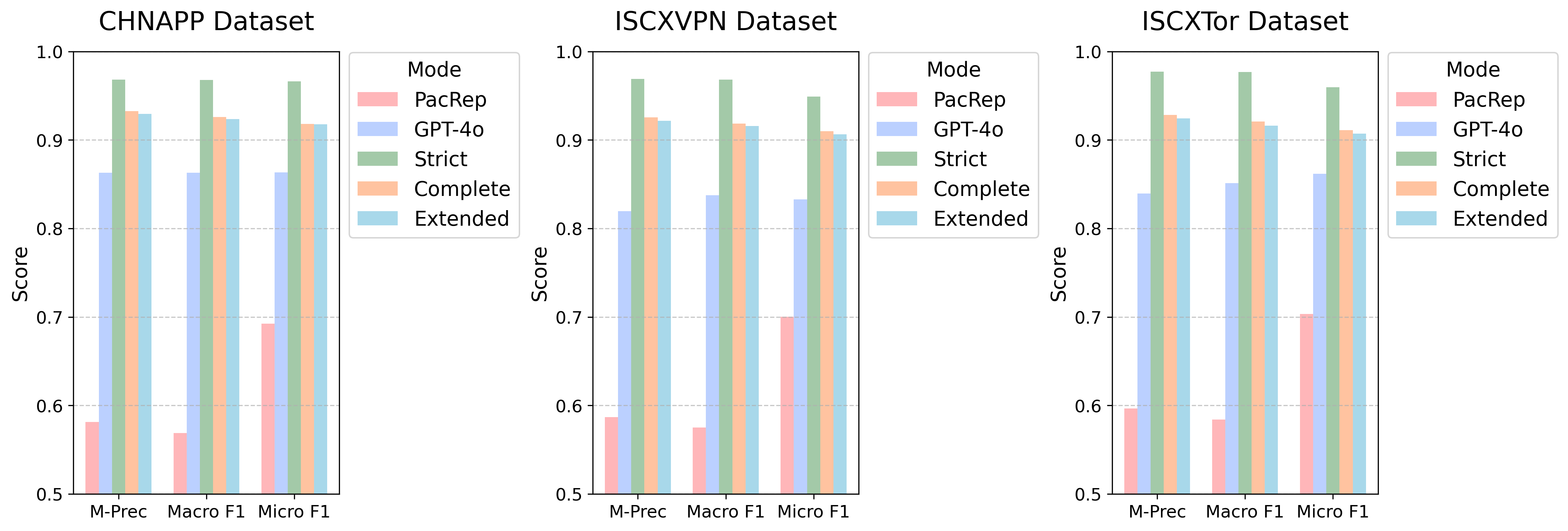}
\caption{Performance comparison of SPS modes (Strict, Complete, Extended) with PacRep-baseline and GPT-4o-baseline across three datasets. For each dataset we report three metrics (M-Prec, Macro F1, Micro F1). Strict mode consistently outperforms both baselines and the other modes, demonstrating the effectiveness of constrained generation space in LLM-based traffic classification.}
\label{fig:prompt_analysis}
\end{figure*}

\textbf{5) Dataset-Specific Performance Analysis:}
Each dataset presents unique challenges that highlight TAO-Net's advantages. The CHNAPP dataset shows the most dramatic improvement in OOD classification accuracy, where TAO-Net reduces misclassification rates by 92.38\% compared to PacRep and 76.72\% compared to GPT-4o, particularly for WeChat and Weibo traffic. The ISCXVPN dataset demonstrates TAO-Net's capability in handling encrypted traffic, where the model maintains clear classification boundaries even under VPN encryption, with minimal confusion between similar service types. The ISCXTor dataset highlights the framework's robustness in multi-layer encrypted scenarios, achieving consistent high accuracy across all categories and showing particular strength in distinguishing between similar application types that confused other models.

\subsection{SPS Analysis} \label{sec:sps_analysis}
In our LLM-based classification scheme, prompts serve as the primary mechanism to shape the large language model's generative space. As the LLM does not merely assign a label but generates a label for network traffic classification, how we formulate each prompt can strongly bias the model's output distribution. Our Semantic-enhanced Prompt Strategy (SPS) explores three distinct modes---Strict, Complete, and Extended---each corresponding to a different level of label diversity provided to the LLM. Table \ref{tab:prompt_comparison_llm} shows prompt templates for these modes, and Figure \ref{fig:prompt_analysis} presents a comprehensive performance comparison across different datasets and metrics, where all SPS modes significantly outperform baselines (PacRep: 58.13-59.64\%, GPT-4o: 81.93-86.30\%), with Strict mode achieving the best results.

\noindent\textbf{Strict Mode.} 
By supplying model with a narrowly constrained set of categories and minimal additional information, Strict mode tightly restricts generative search space. Figure \ref{fig:prompt_analysis} shows this approach consistently achieves superior performance across all metrics and datasets. The CHNAPP dataset demonstrates impressive results with Macro Precision of 96.81\%, Macro F1 of 96.77\%, and Micro F1 of 96.62\%. Similar patterns are observed in ISCXVPN (96.90\%, 96.83\%, 94.91\%) and ISCXTor (97.70\%, 97.68\%, 95.97\%) datasets. This consistent high performance indicates constraining generative space effectively reduces model uncertainty and improves classification precision. However, while such focused prompts can bolster short-term accuracy, they offer limited flexibility in identifying novel or unexpectedly nuanced traffic patterns that lie outside small set of predefined labels.

\noindent\textbf{Complete Mode.} 
In contrast, Complete mode enumerates all dataset-specific labels. This expansion of the prompt naturally broadens the generation space, as the LLM encounters multiple related or overlapping categories when generating predictions. The performance impact of this broader context is evident in Figure \ref{fig:prompt_analysis}, where we observe moderate decreases across metrics. For instance, in the CHNAPP dataset, Complete mode achieves Macro Precision of 93.24\%, Macro F1 of 92.57\%, and Micro F1 of 91.81\%, representing drops of approximately 3-4\% compared to Strict mode. Similar patterns emerge in ISCXVPN (92.53\%, 91.83\%, 91.00\%) and ISCXTor (92.84\%, 92.08\%, 91.10\%) datasets. Our closer inspection indicates that LLM occasionally produces near-synonymous or ambiguous label outputs when two classes share very similar features. Consequently, while coverage improves, classification precision can decline unless further guided by carefully tuned prompt instructions.

\noindent\textbf{Extended Mode.} To push flexibility even further, we inject cross-dataset applications (e.g., Gmail, Skype) into the same prompt. This Extended mode allows LLM to tap into a richer set of applications and potentially recognize traffic patterns that partially overlap between datasets or appear as "new." However, as evidenced by Figure \ref{fig:prompt_analysis}, this increased flexibility comes at a cost to performance. The CHNAPP dataset shows Macro Precision of 92.93\%, Macro F1 of 92.34\%, and Micro F1 of 91.76\%, with similar patterns in ISCXVPN (92.14\%, 91.56\%, 90.64\%) and ISCXTor (92.43\%, 91.62\%, 90.71\%). The performance degradation suggests that expanding the generation space raises the likelihood of confusion when multiple categories exhibit intersecting traffic signatures.

\noindent\textbf{Overall Impact.} 
Our experimental outcomes reinforce a key insight: prompt design plays a crucial role in LLM-based generative classification tasks, with different modes suited to different scenarios. The quantitative results demonstrate that Strict mode outperforms other approaches across all metrics and datasets, with performance gaps ranging from 3 to 7 percentage points. This advantage is particularly evident in complex scenarios - achieving 96.90\% Macro Precision on ISCXVPN (compared to 92.53\% for Complete mode) and 97.70\% on ISCXTor (compared to 92.84\% for Complete mode). This suggests that constraining the generative space can significantly improve classification accuracy when category boundaries are well-defined. However, the trade-off between accuracy and flexibility becomes apparent when considering real-world applications where new traffic patterns may emerge. Complete and Extended modes, while showing lower absolute performance, offer greater adaptability and broader coverage that might be valuable in dynamic network environments. Future research can explore ways to optimize prompt design to achieve a better balance between accuracy and generalization through adaptive prompting strategies that adjust based on traffic characteristics.

\section{Conclusion}

In this paper, we propose a Two-stage Adaptive OOD classification Network (TAO-Net) to address challenges in encrypted traffic classification, particularly for emerging Out-of-Distribution (OOD) traffic. By combining a hybrid OOD detection mechanism with Large Language Models (LLMs) and a Semantic-enhanced Prompt Strategy (SPS), TAO-Net effectively distinguishes between In-Distribution (ID) and OOD traffic. While ensuring accurate classification for known patterns, it also enables fine-grained identification of unseen traffic types without relying on predefined labels. Our approach advances the state-of-the-art by precisely classifying both ID and OOD traffic, excelling at recognizing new network applications. The two-stage design allows TAO-Net to process both ID and OOD traffic, enhancing the detection of novel and potentially malicious applications. Future work will focus on developing an adaptive SPS mechanism and optimizing the OOD detection threshold for different traffic scenarios, further enhancing TAO-Net's applicability in dynamic network environments.

\printcredits
\bibliographystyle{cas-model2-names}
\bibliography{cas-refs}
\end{document}